\documentclass[journal]{IEEEtran}
\usepackage{ragged2e}
\usepackage{times}
\usepackage{epsfig}
\usepackage{graphicx}
\usepackage{amsmath}
\usepackage{amssymb}
\usepackage{multirow}
\usepackage{colortbl}
\usepackage{listings}
\usepackage{footnote}
\usepackage{bm}
\usepackage{cite}
\usepackage{booktabs}
\usepackage{url} 
\usepackage{algorithm}
\usepackage{algorithmic}
\usepackage{colortbl}
\usepackage{hyperref}
\usepackage[dvipsnames]{xcolor}
\usepackage{pifont}
\usepackage{tikz-cd}
\usepackage{microtype}
\usepackage{graphicx}
\usepackage{subcaption}
\usepackage{booktabs} % for professional tables
\usepackage{makecell}
\usepackage{multirow}
\usepackage{booktabs}
\usepackage[dvipsnames]{xcolor}
\usepackage{colortbl}
\definecolor{lightgraycell}{gray}{0.85}
\definecolor{myred}{RGB}{220, 20, 60}
\definecolor{mygreen}{RGB}{0, 128, 0}
\usepackage{siunitx}
\usepackage{makecell}
\usepackage{colortbl} % 如果你要用 \cellcolor
\begin{document}
\title{InfoTok: Information-Theoretic Regularization for Capacity-Constrained Shared Visual Tokenization in Unified MLLMs}

\author{Lv Tang, Tianyi Zheng, Bo Li~\IEEEmembership{Member,~IEEE}, Xingyu Li~\IEEEmembership{Member,~IEEE} \thanks{Lv Tang and Xingyu Li are with University of Alberta, Department of Electrical and Computer Engineering, Edmonton, Canada. Bo Li and Tianyi Zheng are with vivo Mobile Communication Co., Ltd, Shanghai, China. E-mails: luckybird1994@gmail.com, xingyu@ualberta.ca, tyzheng@sjtu.edu.cn, libra@vivo.com.}}

%\markboth{Submit to IEEE Transactions on Image Processing}%

\IEEEtitleabstractindextext{
\begin{abstract}
\justifying
Unified multimodal large language models (MLLMs) aim to unify image understanding and image generation within a single framework, where a shared visual tokenizer serves as the sole interface that maps high-dimensional images into a limited token budget for downstream multimodal reasoning and synthesis.
However, existing shared-token designs are largely architecture-driven and lack an explicit criterion for what information should be preserved to simultaneously support semantic abstraction and visual detail.
In this paper, we adopt a capacity-constrained perspective, viewing the shared tokenizer as a compute-bounded learner whose finite representational budget should prioritize reusable structure over hard-to-exploit high-entropy variations and redundancy.
Motivated by this view, we propose \textbf{\textit{InfoTok}}, an information-regularized tokenization mechanism grounded in the Information Bottleneck (IB) principle.
InfoTok explicitly controls information flow from images to shared tokens to multimodal outputs by imposing mutual-information (MI) constraints that enforce a principled trade-off between compression and task relevance, while also encouraging cross-modal consistency.
Because MI is intractable for high-dimensional visual representations, we instantiate InfoTok with practical, differentiable dependence estimators, including a variational IB formulation and a Hilbert Schmidt Independence Criterion (HSIC) based alternative.
Integrated into three representative unified MLLMs without introducing any additional training data, InfoTok consistently improves both image understanding and generation performance.
These results support information-regularized visual tokenization as a sound basis for token learning in unified MLLMs.
\end{abstract}
\begin{IEEEkeywords}
Unified MLLMs, Image Understanding and Generation, Information-theoretic Regularization
\end{IEEEkeywords}}

\maketitle
\IEEEdisplaynontitleabstractindextext
\IEEEpeerreviewmaketitle

\section{Introduction} \label{sec:intro}

\IEEEPARstart{O}{ver} the past two years, unified multimodal large language models (MLLMs) have progressed rapidly in both image understanding and image generation, with the shared ambition of integrating perception and synthesis within a single framework~\cite{DBLP:conf/cvpr/WuCWMLPLXYR025,DBLP:journals/corr/abs-2505-14683,DBLP:conf/iclr/WuZCTLFZXYY0025,DBLP:journals/corr/abs-2503-21979,huang2025ming,DBLP:journals/corr/abs-2502-20321,DBLP:journals/corr/abs-2506-18898,DBLP:journals/corr/abs-2505-23661,DBLP:journals/corr/abs-2506-03147}.
A central component of these models is the \textbf{\textit{visual tokenizer}}, which converts a high dimensional image into a token sequence that a large language model (LLM) can process.
However, understanding and generation impose inherently different requirements on visual representations.
Understanding favors semantic abstraction, while generation depends on fine grained perceptual detail.
This discrepancy has led to two dominant strategies, decoupled tokenization and shared tokenization, whose evolution reflects the field's pursuit of representational balance in unified modeling.

\begin{figure*}[!htbp]
    \centering
    \includegraphics[width=\linewidth]{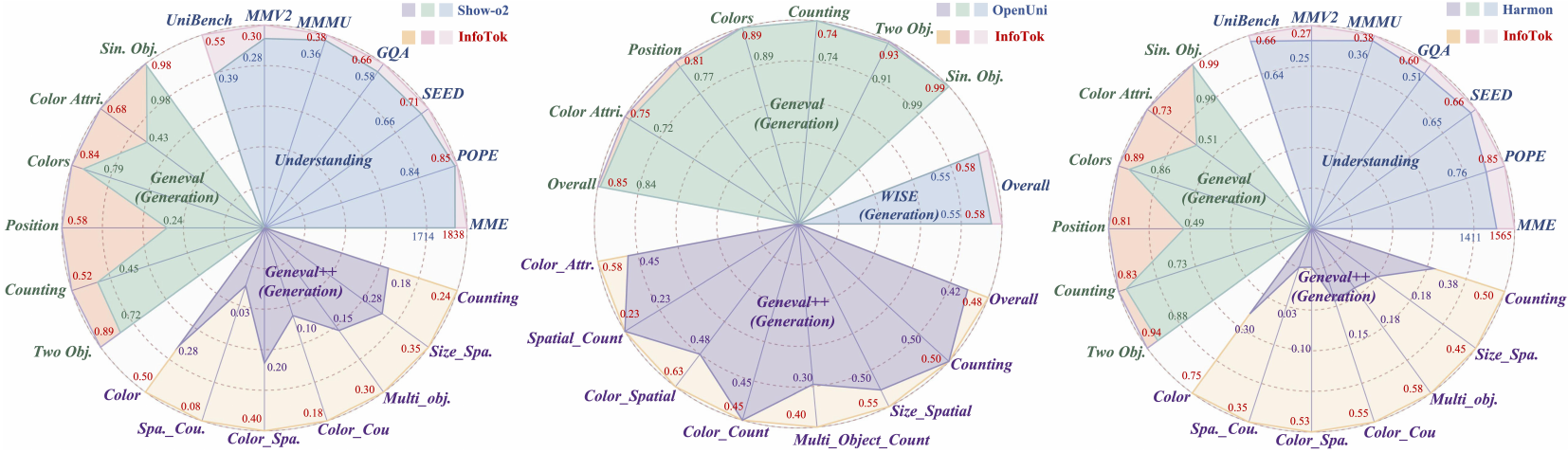}
    \caption{Performance comparison of three representative unified MLLMs (Show-o2~\cite{xie2025show}, OpenUni~\cite{DBLP:journals/corr/abs-2505-23661} and Harmon~\cite{DBLP:journals/corr/abs-2503-21979}) before and after applying InfoTok regularization.
Each radar chart reports results across representative image understanding benchmarks (UniBench~\cite{DBLP:journals/corr/abs-2505-10483}, GQA~\cite{DBLP:conf/cvpr/HudsonM19}, SEED~\cite{DBLP:journals/corr/abs-2307-16125}, POPE~\cite{DBLP:conf/emnlp/LiDZWZW23}, MME~\cite{DBLP:journals/corr/abs-2306-13394}, MMV2~\cite{DBLP:conf/icml/YuYLWL0WW24} and MMMU~\cite{yue2024mmmu}) and image generation benchmarks (Geneval~\cite{DBLP:conf/nips/GhoshHS23}, Geneval++~\cite{DBLP:journals/corr/abs-2508-09987} and WISE~\cite{DBLP:journals/corr/abs-2503-07265}).
The InfoTok-regularized variants achieve consistently improved balance between semantic understanding and visual synthesis, demonstrating that information-theoretic regularization enhances representation sufficiency and cross-modal generalization in unified multimodal learning. All InfoTok fine-tuning experiments are conducted without introducing additional datasets, relying solely on the original training data used by each baseline, indicating that the observed gains primarily stem from the regularized visual tokens learned under our information constraints.}
\vspace{-0.5cm}
    \label{fig_introd1}
\end{figure*}

Recent typical models like Janus~\cite{DBLP:conf/cvpr/WuCWMLPLXYR025} and BAGEL~\cite{DBLP:journals/corr/abs-2505-14683} adopt decoupled visual encoders, using separate branches for understanding and generation.
This separation mitigates cross-task interference by allowing each branch to specialize at an appropriate level of representational granularity.
However, it departs from the goal of a truly shared representation and increases architectural complexity.
Shared-token models construct a single token space for both tasks~\cite{DBLP:conf/iclr/WuZCTLFZXYY0025,DBLP:journals/corr/abs-2503-21979,huang2025ming,DBLP:journals/corr/abs-2502-20321,DBLP:journals/corr/abs-2506-18898}, for example by enlarging codebooks~\cite{DBLP:journals/corr/abs-2502-20321} or aligning visual tokens with text embeddings~\cite{DBLP:conf/iclr/WuZCTLFZXYY0025}.
Despite their progress, existing shared-token methods remain largely architecture-driven.
They rely on end-to-end task losses without an explicit criterion that specifies what information a shared visual token should preserve, leaving the notion of sufficiency for both image understanding and generation tasks unformalized.

Importantly, in shared-token unified MLLMs, the visual tokenizer operates under finite capacity and limited compute when compressing rich visual inputs into a shared token space, and thus behaves as a compute-bounded learner.
Under this constraint, not all visual information is equally useful. Specifically, many low-level details may behave like hard-to-exploit high-entropy variations that waste representational budget, whereas transferable capability is more closely tied to reusable structure such as semantic entities and compositional cues.
We term this observation the capacity-constrained perspective.
It motivates an explicit mechanism to regulate what shared tokens encode, shifting visual tokenization from indiscriminate compression to allocating representational budget toward reusable and valuable information that the compute-bounded visual tokenizer in shared-token unified MLLMs can reliably exploit.

To translate above goal into a concrete objective, we propose \textbf{\textit{InfoTok}}, an information-regularized visual tokenization framework for unified multimodal understanding and generation tasks.
InfoTok treats shared tokenization as an explicit allocation problem under a limited representational budget, controlling the information flow from images to tokens to multimodal outputs so that the resulting tokens remain sufficient for both semantic abstraction and perceptual quality.
Realizing this goal requires a computable criterion to regulate the trade-off, because otherwise the shared token space is determined implicitly by optimization dynamics and the balance between abstraction and fidelity becomes difficult to control.
This motivates an information-theoretic formulation of shared tokenization.
We adopt the Information Bottleneck (IB) principle~\cite{DBLP:journals/corr/physics-0004057,DBLP:conf/itw/TishbyZ15} to formalize InfoTok as an information-regularized learning process, where the tokenization objective is expressed as a trade-off between compression and task relevance through mutual information terms.
However, computing these mutual information terms in closed form is generally infeasible in our setting, since the visual input and discrete tokens lie in heterogeneous spaces and the underlying distributions are inaccessible.
To obtain a computable objective, we instantiate InfoTok with tractable dependence estimators, adopting variational IB (VIB) ~\cite{DBLP:conf/iclr/AlemiFD017} or Hilbert Schmidt Independence Criterion (HSIC)~\cite{DBLP:conf/nips/WangJMID21} based dependence measure as two feasible instantiations that serve the same role of regulating information dependencies during training. As a result, our proposed InfoTok promotes compact visual tokens that preserve reusable semantic and compositional structure, together with task-relevant perceptual cues.

We integrate InfoTok into three representative unified MLLMs, including Harmon~\cite{DBLP:journals/corr/abs-2503-21979}, OpenUni~\cite{DBLP:journals/corr/abs-2505-23661}, and Show-o2~\cite{xie2025show}, and fine tune them without additional training data.
As shown in Fig.~\ref{fig_introd1}, InfoTok consistently enhances both understanding and generation performance, supporting our thesis that explicitly regulating information flow yields a more optimal and stable shared token space.
Beyond empirical gains, we provide a theoretical analysis that further supports information regularized tokenization as a coherent foundation for shared token learning, and offers insight into the feasibility of unifying multimodal understanding and generation within a single representation.
From our empirical results and theoretical analysis, we suggest that, beyond architectural scaling, principled information control in shared visual tokens can serve as a general design tool for improving the stability and performance of unified MLLMs on both understanding and generation.
Contributions of this work are summarized as:

\begin{itemize}
    
    \item We introduce a capacity-constrained perspective for shared tokenization, highlighting that in shared-token unified MLLMs the visual tokenizer operates as a compute-bounded learner under limited representational capacity and computation, and thus requires an explicit criterion to allocate budget toward reusable structure rather than high-entropy visual variations and redundancy.
    
    \item We ground shared token learning in an information-theoretic objective. Building on IB principle, we formalize InfoTok by modeling the information dependencies among visual inputs, shared tokens, and multimodal outputs, which yields an explicit regularization objective for shared visual tokenization. To make this objective computable, we derive tractable instantiations using VIB and HSIC.

    \item We integrate InfoTok into three representative unified MLLMs and fine tune them without introducing additional training data. Experiments demonstrate consistent improvements on both understanding and generation, indicating that explicit information regulation leads to a more effective shared token space. We further provide theoretical analysis that supports the rationale of information regularized tokenization in the unified setting. We hope findings in this paper can suggest a practical path for translating the theory into the design of unified MLLMs.
    
\end{itemize}

\section{Related Work}
The rapid advancement of MLLMs has opened a new frontier in artificial intelligence, driving increasing interest in unified architectures that can both understand and generate multimodal content~\cite{yin2024survey,DBLP:journals/csur/YuanLZ25}.
This section reviews the evolution from specialized multimodal systems to unified frameworks, highlighting visual tokenization and the resulting tension between semantic abstraction for understanding and appearance cues for generation that motivates our proposed InfoTok.

\subsection{Multimodal Large Language Models}

\noindent \textbf{Image Understanding.}
Understanding oriented MLLMs aim to interpret visual inputs and produce textual outputs, spanning tasks such as visual question answering, image captioning, and general visual reasoning.
A common architectural pattern connects a pretrained vision encoder to a pretrained LLM through a lightweight interface, enabling next token prediction over interleaved visual and textual tokens.
Representative models include LLaVA~\cite{DBLP:conf/nips/LiuLWL23a} and MiniGPT-4~\cite{DBLP:conf/iclr/Zhu0SLE24}, as well as a broader family of instruction tuned models~\cite{DBLP:conf/nips/Dai0LTZW0FH23,DBLP:conf/icml/0008LSH23}.
In these architectures, the interface is not merely an implementation detail, but the key component that determines how continuous visual features are transformed into a compact set of tokens compatible with the LLM.
Accordingly, existing designs can be broadly grouped into two categories.
One line uses projector based mappings that directly align vision encoder features with the language embedding space~\cite{DBLP:conf/nips/LiuLWL23a,DBLP:conf/iclr/Zhu0SLE24}.
The other line adopts query based resampling modules that select a compact set of visual tokens from dense visual features, improving efficiency and reducing redundancy~\cite{alayrac2022flamingo,liang2024querying}.

\noindent \textbf{Image Generation.}
Generation oriented models focus on synthesizing images from textual prompts and have undergone evolution in both modeling and representation.
Diffusion based approaches~\cite{DBLP:conf/cvpr/RombachBLEO22,DBLP:conf/iccv/PeeblesX23,DBLP:journals/corr/abs-2303-07909,DBLP:conf/cvpr/ZhangHLG025,DBLP:journals/corr/abs-2204-06125} formulate synthesis as iterative denoising and achieve strong fidelity, but their generation process is typically not expressed as standard next token prediction.
A complementary direction explores autoregressive generation in a discrete token space, where images are represented as sequences of visual tokens and generated in the same next token prediction form as language.
This direction relies critically on image tokenization and discrete representation learning, since the quality and efficiency of generation are closely tied to the fidelity and compactness of the token space.
More recent studies further move toward fully autoregressive multimodal modeling, in which text and image tokens are concatenated and generated under a unified next token prediction objective~\cite{DBLP:conf/cvpr/MuV025,DBLP:conf/iclr/DongHPQGYZSZWK024,DBLP:conf/iclr/GeZZGLWS24,DBLP:conf/cvpr/SunCZZYWRL0W24,DBLP:journals/corr/abs-2405-09818}.
This formulation lays the groundwork for unified multimodal frameworks that aim to support both understanding and generation within a single model~\cite{DBLP:journals/corr/abs-2504-18391}.

\subsection{Unified Multimodal Large Language Models}

Building on autoregressive multimodal modeling, unified MLLMs aim to support both understanding and generation within a single architecture~\cite{DBLP:conf/cvpr/WuCWMLPLXYR025,DBLP:journals/corr/abs-2505-14683,DBLP:conf/iclr/WuZCTLFZXYY0025,DBLP:journals/corr/abs-2503-21979,huang2025ming,DBLP:journals/corr/abs-2502-20321,DBLP:journals/corr/abs-2506-18898,DBLP:journals/corr/abs-2505-23661,DBLP:journals/corr/abs-2506-03147,DBLP:journals/corr/abs-2409-18869,DBLP:journals/corr/abs-2411-17762,wang2025growing,DBLP:journals/corr/abs-2505-05422,DBLP:journals/corr/abs-2505-09568}.
A central design question is how to represent images as tokens in a way that simultaneously supports semantic understanding and high fidelity synthesis.
In recent years, two major tokenization paradigms have emerged.

\noindent \textbf{Decoupled tokenization.}
Decoupled tokenization frameworks employ separate encoders or tokenizers for understanding and generation, allowing each branch to specialize in semantic abstraction or perceptual detail, as in Janus~\cite{DBLP:conf/cvpr/WuCWMLPLXYR025} and BAGEL~\cite{DBLP:journals/corr/abs-2505-14683}.
While this separation can reduce cross task interference, it introduces heavier pipelines and fragmented representation spaces, and it departs from the goal of unification by construction.

\noindent \textbf{Shared tokenization.}
Shared tokenization frameworks pursue a single visual code space used by both understanding and generation, as exemplified by VILA-U~\cite{DBLP:conf/iclr/WuZCTLFZXYY0025}, Harmon~\cite{DBLP:journals/corr/abs-2503-21979}, and UniTok~\cite{DBLP:journals/corr/abs-2505-23661}.
Although these methods differ in implementation, they can be broadly organized by how they attempt to reconcile abstraction and fidelity within a shared token space.
A first line emphasizes alignment oriented designs, introducing alignment priors that encourage visual tokens to become more compatible with language embeddings~\cite{DBLP:conf/iclr/WuZCTLFZXYY0025}.
A second line leverages objective oriented pretexts that reshape the token space through masked prediction or multimodal reconstruction style training, aiming to support both interpretation and synthesis~\cite{DBLP:journals/corr/abs-2503-21979}.
A third line scales quantization capacity, expanding codebooks or token budgets to accommodate both semantic cues and fine details~\cite{DBLP:journals/corr/abs-2505-23661,DBLP:journals/corr/abs-2502-20321}.
Despite their progress, shared tokenization approaches remain largely architecture driven and empirically tuned.
Most rely on design choices and training dynamics to shape the information content of shared tokens, but they lack an explicit principle that governs what information should be retained and how the information trade off should be controlled across tasks.
Our work addresses this limitation by introducing InfoTok, which grounds shared token learning in the IB principle~\cite{DBLP:journals/corr/physics-0004057,DBLP:conf/itw/TishbyZ15}.
Moreover, we derive practical instantiations that model the dependencies among visual inputs, tokenized representations, and multimodal outputs, enabling controllable information retention and providing a principled path toward balanced unified multimodal representation learning.

\begin{figure*}[!htbp]
    \centering
    \includegraphics[width=\linewidth]{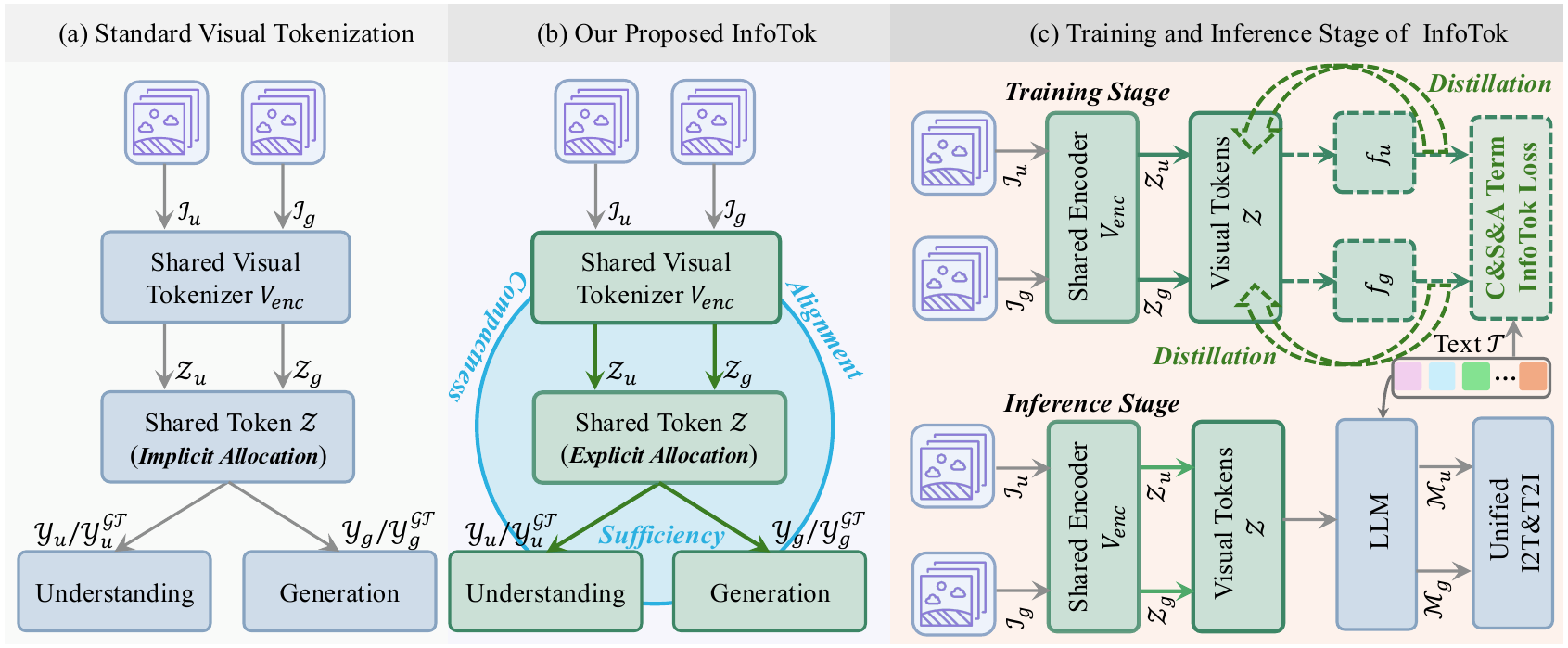}
    \caption{Illustration of our information-regularized tokenization (InfoTok).
    (a) depicts a standard unified MLLM with shared tokenization, where a single visual tokenizer jointly supports understanding (I2T) and generation (T2I) tasks.
    (b) presents our proposed InfoTok, which imposes a principled information-regularization objective to explicitly control what visual content is retained in the shared token space. Specifically, InfoTok encourages compact tokens by suppressing redundant input information under a limited token budget (\textbf{\textit{Compactness}}), while ensuring that the remaining tokens still preserve the task-critical semantics for understanding and the fine-grained perceptual cues needed for generation (\textbf{\textit{Sufficiency)}}.
    (c) illustrates the training and inference workflows of InfoTok. During training, the compactness, sufficiency, and alignment objectives, together with the distillation term, form the InfoTok loss to regularize the shared visual tokens in existing unified MLLMs.}
    \label{framework}
    \vspace{-0.5cm}
\end{figure*}

\section{Proposed Method}
In this section, we first outline the standard unified MLLM architecture.
We then introduce InfoTok and reformulate shared visual tokenization through the IB principle.
This leads to an explicit information-regularized objective that governs token formation.
Finally, we derive a tractable training objective with practical instantiations via variational IB and HSIC, which can be applied to regularize existing unified MLLMs.

\subsection{Unified MLLM Frameworks} \label{sec_unified_mllm}

A typical unified MLLM processes multimodal inputs consisting of images \(\mathcal{I}=\{\mathcal{I}_{u}, \mathcal{I}_{g}\}\) and texts \(\mathcal{P}=\{\mathcal{P}_{u}, \mathcal{P}_{g}\}\), where subscripts \(u\) and \(g\) denote understanding and generation respectively. A typical unified MLLM operates through following three stages:

\noindent\textbf{1) Visual and Text Tokenization.}
The shared visual encoder $V_{enc}$ maps the input image $\mathcal{I}$ into a sequence of visual tokens \(\mathcal{Z}=\{\mathcal{Z}_{u}, \mathcal{Z}_{g}\}\), while a text tokenizer converts the text prompt \(\mathcal{P}\) into a sequence of text tokens \(\mathcal{T}=\{\mathcal{T}_{u}, \mathcal{T}_{g}\}\).

\noindent\textbf{2) Multimodal Fusion.}
The LLM (e.g., Qwen2.5~\cite{DBLP:journals/corr/abs-2412-15115}, Ling-lite~\cite{DBLP:journals/corr/abs-2503-05139} and LLaMA~\cite{touvron2023llama}) receives the concatenated text and visual tokens and performs the next-token prediction.
During training, the model learns to predict either the next text token \(\mathcal{M}_u\) or the next visual token \(\mathcal{M}_g\), thereby unifying image understanding and generation tasks within a single objective.

\noindent\textbf{3) De-Tokenization.}
The predicted tokens are decoded into the target modality using a text decoder for \(\mathcal{M}_u\) and a visual decoder for \(\mathcal{M}_g\), producing the final outputs \(\mathcal{Y}=\{\mathcal{Y}_{u}, \mathcal{Y}_{g}\}\).

As a result, the overall training objective of a typical unified MLLM is formulated as the combination of text-to-image (T2I) and image-to-text (I2T) losses:
\begin{equation}
\mathcal{L}_{\text{MLLM}} =
\mathcal{L}_{\text{t2i}}(\mathcal{Y}_g, \mathcal{Y}_g^{\mathcal{GT}}) +
\mathcal{L}_{\text{i2t}}(\mathcal{Y}_u, \mathcal{Y}_u^{\mathcal{GT}}),
\label{Eqn_MLLM_Training}
\end{equation}
where \(\mathcal{L}_{\text{i2t}}\) typically denotes a cross-entropy loss for text generation ~\cite{DBLP:conf/iclr/XieMBZWLGCYS25,DBLP:journals/corr/abs-2501-17811}. \(\mathcal{L}_{\text{t2i}}\) typically represents a diffusion-based loss for image generation~\cite{DBLP:conf/iclr/ZhouYBTYSKMZL25,DBLP:journals/corr/abs-2505-14683}.
This objective enables unified end-to-end optimization of both understanding and generation tasks within a multimodal framework.

In the typical unified MLLM framework, the latent visual representation $\mathcal{Z}$ acts as the bridge between the visual encoder and the LLM backbone, serving as the sole conduit of visual information.
The informativeness and structure of $\mathcal{Z}$ are therefore pivotal in determining the model’s ability to achieve semantic understanding and high-quality visual generation.

\subsection{Unified Information-Regularized Tokenization} \label{sec_uni_infotok}
This subsection formulates tokenization from an information-theoretic viewpoint, with the goal of explicitly regulating what information is retained in the visual representation $\mathcal{Z}$.
We first review the classical IB principle, then extend it to derive an objective that encourages $\mathcal{Z}$ to be compact while remaining sufficient for unified multimodal understanding and generation.

\subsubsection{Classical IB Formulation}
From information-theoretic perspective, representation \(\mathcal{Z}\) learned by a multimodal model can be viewed as IB formulation~\cite{DBLP:journals/corr/physics-0004057,DBLP:conf/itw/TishbyZ15}.
The IB framework posits that an optimal representation should capture only the information from the input that is relevant to predicting the output while discarding irrelevant information.

Formally, given the input image $\mathcal{I}$ and the target $\mathcal{Y}^\mathcal{GT}$, 
the IB principle seeks to learn a visual latent representation $\mathcal{Z}$ by finding the infimum of the following Lagrangian:
\begin{equation}
\inf_{p(\mathcal{Z}|\mathcal{I})} \mathcal{L}_{\text{IB}} 
\quad \text{where} \quad 
\mathcal{L}_{\text{IB}} = I(\mathcal{Z}; \mathcal{I}) - \beta I(\mathcal{Z};{\mathcal{Y}^\mathcal{GT}}).
\end{equation}
Herein, $I(\cdot; \cdot)$ denotes mutual information, and $\beta \ge 0$ is the Lagrangian multiplier balancing the trade-off 
between information constraint $I(\mathcal{Z}; \mathcal{I})$ and predictive sufficiency $I(\mathcal{Z};{\mathcal{Y}^\mathcal{GT}})$. 
The conditional distribution $p(\mathcal{Z}|\mathcal{I})$ represents the mapping modeled by the shared visual encoder, describing how the latent representation $\mathcal{Z}$ is stochastically generated from the input image $\mathcal{I}$ 
under the information bottleneck formulation. 
It formalizes how the shared visual encoder $V_{enc}$ mediates information retention and abstraction, 
acting as a probabilistic mapping from visual inputs $\mathcal{I}$ to latent visual tokens $\mathcal{Z}$.

\begin{figure*}[!htbp]
    \centering
    \includegraphics[width=\linewidth]{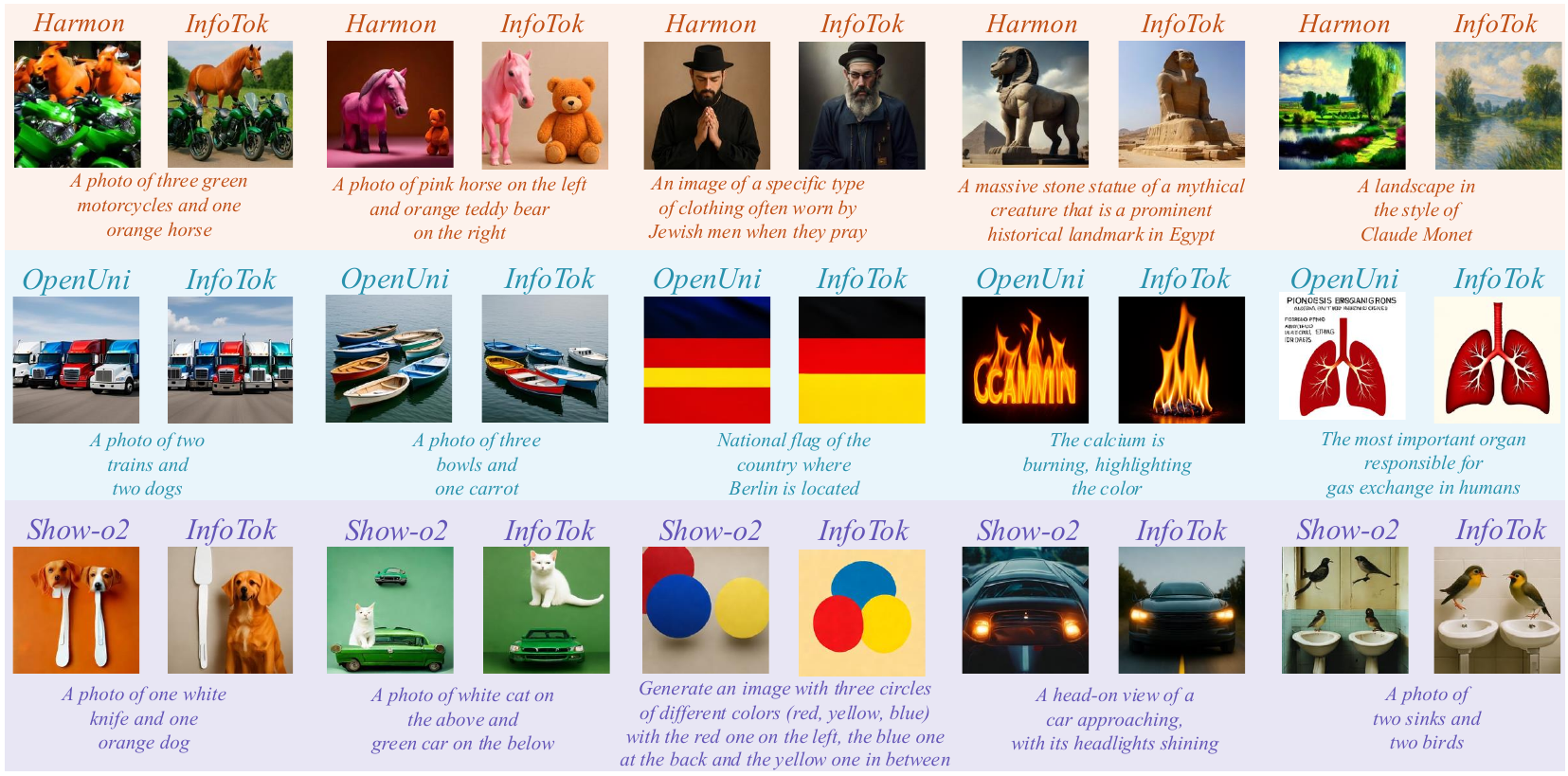}
    \caption{We apply InfoTok to three representative shared-token unified MLLMs and observe consistent improvements in generation performance. From the qualitative visualizations, the gains are clearly evident: with InfoTok, the models show stronger instruction following (the first two Show-o2 samples) and higher overall generation quality (the last samples of Harmon and OpenUni). These results suggest that InfoTok encourages the shared tokens to retain more task-relevant information while suppressing redundant content. Notably, all InfoTok fine-tuning experiments are conducted without additional datasets, relying solely on each baseline’s original training data, which highlights the effectiveness of the proposed information-regularized tokenization.}
    \label{fig_visualisation}
    \vspace{-0.5cm}
\end{figure*}

\subsubsection{Extended IB Formulation in InfoTok}

The classical IB principle offers a principled objective for learning a representation that trades off compression and task relevance for a single input--output mapping.
In unified MLLMs, a shared visual tokenizer must support two distinct objectives, understanding and generation, and remain compatible with textual tokens in the unified next-token prediction pipeline.
Under the unified MLLM definition in Section~\ref{sec_unified_mllm}, the shared visual encoder produces task-conditioned visual tokens defined as follows:
\begin{equation}
\mathcal{Z}_u = V_{\text{enc}}(\mathcal{I}_u), 
\quad
\mathcal{Z}_g = V_{\text{enc}}(\mathcal{I}_g).
\label{Eqn_Z_from_encoder}
\end{equation}

To make the task requirements explicit when formulating information constraints, we introduce lightweight task-specific MLP projections on top of these encoder tokens during training:
\begin{equation}
\tilde{\mathcal{Z}}_u = f_u(\mathcal{Z}_u), 
\quad
\tilde{\mathcal{Z}}_g = f_g(\mathcal{Z}_g),
\label{Eqn_Z_tilde_def}
\end{equation}
where $\tilde{\mathcal{Z}}_u$ and $\tilde{\mathcal{Z}}_g$ act as task-facing representations for defining the InfoTok regularization terms.
These projections are used only during training and can be removed at inference.
Importantly, because $f_u$ and $f_g$ are deterministic post-processing of $\mathcal{Z}$, they cannot increase mutual information in the sense of the data processing inequality.
This observation motivates applying information regulation on $\tilde{\mathcal{Z}}$ as a practical handle for shaping the shared encoder, a point that we formalize later.

Based on this structure, InfoTok extends the IB objective to jointly regulate tokenization for understanding and generation while encouraging consistent interaction with textual tokens.
We define the following information-regularized objective:
\begin{equation}
\begin{aligned}
\mathcal{L}_{\text{IB}}^{(u)} &= I(\tilde{\mathcal{Z}}_u; \mathcal{I}_u)
 - \beta_u I(\tilde{\mathcal{Z}}_u; \mathcal{Y}_u^\mathcal{GT})
 - \alpha_u I(\tilde{\mathcal{Z}}_u; \mathcal{T}_u), \\
\mathcal{L}_{\text{IB}}^{(g)} &= I(\tilde{\mathcal{Z}}_g; \mathcal{I}_g)
 - \beta_g I(\tilde{\mathcal{Z}}_g; \mathcal{Y}_g^\mathcal{GT})
 - \alpha_g I(\tilde{\mathcal{Z}}_g; \mathcal{T}_g).
\end{aligned}
\label{Eqn_ExtendedIB}
\end{equation}
Here $\alpha$ and $\beta$ are positive Lagrange multipliers that control the trade-offs among the following objectives.

\noindent \textbf{1) Compactness:}
The term $I(\tilde{\mathcal{Z}};\mathcal{I})$ penalizes redundant input information in the projected tokens, promoting compact representations under a limited token budget.

\noindent \textbf{2) Sufficiency:}
The term $I(\tilde{\mathcal{Z}};\mathcal{Y}^{\mathcal{GT}})$ encourages the tokens to retain task-relevant information needed to predict the targets for image understanding and generation.

\noindent \textbf{3) Alignment:}
The term $I(\tilde{\mathcal{Z}};\mathcal{T})$ encourages compatibility between visual and textual representations, reinforcing cross-modal consistency in the unified modeling pipeline.

Finally, the InfoTok objective is defined as follows:
\begin{equation}
\mathcal{L}_{\text{InfoTok}} = \mathcal{L}_{\text{IB}}^{(u)} + \mathcal{L}_{\text{IB}}^{(g)}.
\label{Eqn_InfoTok}
\end{equation}
Although Eqn.~\ref{Eqn_ExtendedIB} is written in terms of the projected tokens $\tilde{\mathcal{Z}}_u$ and $\tilde{\mathcal{Z}}_g$, the regularization is applied to the shared encoder through the compositional structure in Eqn.~\ref{Eqn_Z_from_encoder} and Eqn.~\ref{Eqn_Z_tilde_def}.
In other words, constraining the information content of $\tilde{\mathcal{Z}}$ induces corresponding pressure on $\mathcal{Z}$ and thus on $V_{\text{enc}}$.
Together, these terms provide a unified and principled mechanism for regulating information flow in shared visual tokenization, improving the stability of a single token space for both multimodal understanding and generation tasks.

\subsubsection{Variational IB Estimation in InfoTok}

The extended InfoTok objective in Eqn.~\ref{Eqn_ExtendedIB} is expressed in terms of mutual information quantities, including 
$I(\tilde{\mathcal{Z}};\mathcal{I})$, $I(\tilde{\mathcal{Z}};\mathcal{Y}^{\mathcal{GT}})$, and $I(\tilde{\mathcal{Z}};\mathcal{T})$.
Directly evaluating these terms is generally intractable for high-dimensional visual representations produced by deep neural networks, as it would require access to the underlying joint and marginal distributions (e.g., $p(\tilde{\mathcal{Z}},\mathcal{I})$ and $p(\tilde{\mathcal{Z}})$) that are not available in closed form~\cite{DBLP:conf/icml/PooleOOAT19}.
Therefore, optimizing Eqn.~\ref{Eqn_ExtendedIB} calls for practical dependence estimators that are computable and differentiable during the training stage.

To this end, InfoTok is instantiated with two estimators that play the same role of regulating the information dependencies in Eqn.~\ref{Eqn_ExtendedIB}$:$ 
(i) a Variational Information Bottleneck (VIB) formulation~\cite{DBLP:conf/iclr/AlemiFD017}, and 
(ii) an HSIC-based dependence measure~\cite{DBLP:conf/nips/WangJMID21}.
Following Eqn.~\ref{Eqn_Z_from_encoder} and Eqn.~\ref{Eqn_Z_tilde_def}, we define the task-facing tokens as $\tilde{\mathcal{Z}}=f(\mathcal{Z})$ with $\mathcal{Z}=V_{\text{enc}}(\mathcal{I})$, and model the corresponding tokenization through a variational posterior $q_{\phi}(\tilde{\mathcal{Z}} \mid \mathcal{I})$.

Under the VIB instantiation, the intractable mutual-information terms involving $\tilde{\mathcal{Z}}$ are replaced by variational bounds, yielding differentiable surrogates that enable end-to-end optimization while preserving the information-theoretic interpretation of InfoTok.
Concretely, $q_{\phi}(\tilde{\mathcal{Z}} \mid \mathcal{I})$ is parameterized as a Gaussian whose mean $\mu_{\phi}(\mathcal{I})$ and variance $\sigma_{\phi}^{2}(\mathcal{I})$ are predicted by a neural network, and $\tilde{\mathcal{Z}}$ is sampled via the reparameterization trick from 
$\mathcal{N}(\mu_{\phi}(\mathcal{I}), \mathrm{diag}(\sigma_{\phi}^{2}(\mathcal{I})))$~\cite{DBLP:journals/corr/KingmaW13-vae}.
This probabilistic relaxation leads to tractable estimators for the three components in Eqn.~\ref{Eqn_ExtendedIB}.

In parallel, HSIC provides a kernel-based dependence statistic that does not require explicit density modeling, and can be used as an alternative proxy to regulate the same dependencies between representations and their conditioning variables.
Since VIB and HSIC play the same role in InfoTok, namely making the information constraints in Eqn.~\ref{Eqn_ExtendedIB} computable, we present the VIB-based derivation in the main text for clarity and include HSIC as an alternative instantiation in the experiments.
With these estimators, the compactness term, task sufficiency term, and cross-modal alignment term in InfoTok are defined as:

\noindent\textbf{1) Compactness Term.}
The compactness term penalizes redundant input information in the projected tokens, encouraging the tokenizer to retain only the essential visual content under a limited token budget.
Consistent with Eqn.~\ref{Eqn_ExtendedIB}, we regulate the mutual information between the projected tokens and their corresponding visual inputs.
For the understanding and generation branches, we have the standard VIB upper bounds:
\begin{equation}
\begin{aligned}
I(\tilde{\mathcal{Z}}_u; \mathcal{I}_u) 
&\le \mathbb{E}_{p(\mathcal{I}_u)}\!\left[D_{\mathrm{KL}}\!\big(q_\phi(\tilde{\mathcal{Z}}_u|\mathcal{I}_u)\,\|\,r_u(\tilde{\mathcal{Z}}_u)\big)\right],\\
I(\tilde{\mathcal{Z}}_g; \mathcal{I}_g) 
&\le \mathbb{E}_{p(\mathcal{I}_g)}\!\left[D_{\mathrm{KL}}\!\big(q_\phi(\tilde{\mathcal{Z}}_g|\mathcal{I}_g)\,\|\,r_g(\tilde{\mathcal{Z}}_g)\big)\right],
\end{aligned}
\label{Eqn_IB_Compactness}
\end{equation}
where $r_u(\tilde{\mathcal{Z}}_u)$ and $r_g(\tilde{\mathcal{Z}}_g)$ denote the priors for the two branches, typically chosen as $\mathcal{N}(0,\mathbf{I}_d)$.
Minimizing the KL terms encourages compact projected tokens by filtering redundant visual details while preserving task useful content.

\noindent\textbf{2) Task Sufficiency Term.}
While compactness reduces redundancy, the projected tokens must remain predictive of task specific targets.
Accordingly, InfoTok maximizes $I(\tilde{\mathcal{Z}}_u; \mathcal{Y}_u^{\mathcal{GT}})$ for understanding and $I(\tilde{\mathcal{Z}}_g; \mathcal{Y}_g^{\mathcal{GT}})$ for generation.
By introducing variational decoders $p_\theta(\mathcal{Y}_u^{\mathcal{GT}}|\tilde{\mathcal{Z}}_u)$ and $p_\theta(\mathcal{Y}_g^{\mathcal{GT}}|\tilde{\mathcal{Z}}_g)$, we can obtain tractable lower bounds:
\begin{equation}
\begin{aligned}
I(\tilde{\mathcal{Z}}_u; \mathcal{Y}_u^{\mathcal{GT}})
&\ge \mathbb{E}_{p(\mathcal{I}_u,\mathcal{Y}_u^{\mathcal{GT}})}
\Big[\log p_\theta(\mathcal{Y}_u^{\mathcal{GT}} \mid \tilde{\mathcal{Z}}_u)\Big]
+ H(\mathcal{Y}_u^{\mathcal{GT}}),\\
I(\tilde{\mathcal{Z}}_g; \mathcal{Y}_g^{\mathcal{GT}})
&\ge \mathbb{E}_{p(\mathcal{I}_g,\mathcal{Y}_g^{\mathcal{GT}})}
\Big[\log p_\theta(\mathcal{Y}_g^{\mathcal{GT}} \mid \tilde{\mathcal{Z}}_g)\Big]
+ H(\mathcal{Y}_g^{\mathcal{GT}}),
\end{aligned}
\label{Eqn_Task_MI}
\end{equation}
where $H(\mathcal{Y}^{\mathcal{GT}})$ is constant with respect to optimization.
Maximizing these bounds encourages the projected tokens to preserve task relevant semantic and perceptual information required for both reasoning and synthesis.

\noindent\textbf{3) Alignment Term.}
To promote cross modal consistency in the unified next token prediction, InfoTok maximizes the mutual information between projected visual tokens and their corresponding textual embeddings, namely $I(\tilde{\mathcal{Z}}_u;\mathcal{T}_u)$ and $I(\tilde{\mathcal{Z}}_g;\mathcal{T}_g)$.
Since direct mutual information computation remains intractable, we adopt a contrastive estimator that provides a tractable lower bound.
For each branch, we form an aggregated visual embedding from the posterior mean,
\(\bar{{\mathcal{Z}}}_u = \mathrm{mean}(\mu_\phi(\mathcal{I}_u))\) and
\(\bar{{\mathcal{Z}}}_g = \mathrm{mean}(\mu_\phi(\mathcal{I}_g))\),
and apply an InfoNCE based estimator~\cite{DBLP:journals/corr/abs-1807-03748}:
\begin{equation}
\begin{aligned}
\widehat{I}(\tilde{\mathcal{Z}}_u; \mathcal{T}_u)
&= \mathbb{E}\!\left[
\log
\frac{\exp(\mathrm{sim}(\bar{{\mathcal{Z}}}_u, \mathcal{T}_u)/\tau)}
{\sum_{k}\exp(\mathrm{sim}(\bar{{\mathcal{Z}}}_u, \mathcal{T}_{u,k})/\tau)}
\right],\\
\widehat{I}(\tilde{\mathcal{Z}}_g; \mathcal{T}_g)
&= \mathbb{E}\!\left[
\log
\frac{\exp(\mathrm{sim}(\bar{{\mathcal{Z}}}_g, \mathcal{T}_g)/\tau)}
{\sum_{k}\exp(\mathrm{sim}(\bar{{\mathcal{Z}}}_g, \mathcal{T}_{g,k})/\tau)}
\right],
\end{aligned}
\label{Eqn_InfoNCE_Branch}
\end{equation}
where $\mathrm{sim}(\cdot,\cdot)$ denotes cosine similarity, $\tau$ is a temperature parameter, and $\mathcal{T}_{u,k}$ and $\mathcal{T}_{g,k}$ denote negative textual samples.
These estimators act as contrastive lower bounds of the corresponding mutual information terms, encouraging the visual tokens to become semantically compatible with text and thereby improving cross-modal coherence in the shared space.

\noindent\textbf{Final Objective.}  
Above three terms together approximate the mutual information components in Eqn.~\ref{Eqn_ExtendedIB} 
through tractable variational bounds, with compactness providing an upper bound via KL divergence.
The sufficiency and alignment terms provide tractable lower bounds on their respective mutual information objectives via likelihood maximization and contrastive estimation.
By combining these approximations, our InfoTok can yield a unified, differentiable loss function that integrates all information constraints into a training objective:
\begin{equation}
\label{Eqn_Final_InfoTok_Explicit}
\begin{aligned}
\mathcal{L}_{\text{InfoTok}}^{(u)} 
&= \mathbb{E}_{p(\mathcal{I}_u)} \big[D_{\text{KL}}\!\big(q_\phi(\tilde{\mathcal{Z}}_u|\mathcal{I}_u)\,\|\,r_u(\tilde{\mathcal{Z}}_u)\big)\big] \\
 &- \beta_u \mathbb{E}_{p(\mathcal{I}_u, \mathcal{Y}_u^{\mathcal{GT}})} 
    \big[\log p_\theta(\mathcal{Y}_u^{\mathcal{GT}} | \tilde{\mathcal{Z}}_u)\big]
 - \alpha_u \widehat{I}(\tilde{\mathcal{Z}}_u;\mathcal{T}_u), \\
\mathcal{L}_{\text{InfoTok}}^{(g)} 
&= \mathbb{E}_{p(\mathcal{I}_g)} \big[D_{\text{KL}}\!\big(q_\phi(\tilde{\mathcal{Z}}_g|\mathcal{I}_g)\,\|\,r_g(\tilde{\mathcal{Z}}_g)\big)\big] \\
 & - \beta_g \mathbb{E}_{p(\mathcal{I}_g, \mathcal{Y}_g^{\mathcal{GT}})} 
    \big[\log p_\theta(\mathcal{Y}_g^{\mathcal{GT}} | \tilde{\mathcal{Z}}_g)\big]
 - \alpha_g \widehat{I}(\tilde{\mathcal{Z}}_g; \mathcal{T}_g), \\
 \mathcal{L}_{\text{InfoTok}} &= \mathcal{L}_{\text{InfoTok}}^{(u)} + \mathcal{L}_{\text{InfoTok}}^{(g)}.
\end{aligned}
\end{equation}

\begin{table}[!t]
\centering
\caption{To verify whether InfoTok yields positive effects in unified MLLMs, we apply our information-regularized training to Harmon and Show-o2 and re-train the shared visual tokenizer $V_{\text{enc}}$.
To assess generation, we freeze $V_{\text{enc}}$ and train a lightweight decoder on ImageNet~\cite{DBLP:conf/cvpr/DengDSLL009}, then evaluate reconstruction quality on the ImageNet validation set using FID.
To assess understanding, we measure cross-modal dependence between visual tokens and the final text representations on DenseFusion-1M~\cite{li2024densefusion} using normalized centered kernel alignment (CKA), with 10k randomly sampled images that do not overlap with training data.}
\label{infotok_vae_mi}
\resizebox{\linewidth}{!}{%
\begin{tabular}{lcc|cc}
\toprule
\textbf{Metric} &
\textbf{Harmon} &
\thead{\textbf{InfoTok}\\[-1pt]\textbf{(Harmon)}} &
\textbf{Show-o2} &
\thead{\textbf{InfoTok}\\[-1pt]\textbf{(Show-o2)}} \\
\midrule
FID$\downarrow$  & 14.1 & \textbf{\textcolor{red}{11.8}} & 13.0 & \textbf{\textcolor{red}{11.3}} \\
CKA$\uparrow$    & 0.243 & \textbf{\textcolor{red}{0.274}} & 0.236 & \textbf{\textcolor{red}{0.285}} \\
\bottomrule
\end{tabular}}
\vspace{-0.5cm}
\end{table}

\begin{table*}[t]
\centering
\caption{Comparison of unified MLLMs on Geneval and WISE. Gray-shaded InfoTok rows show absolute improvements.
\textbf{Boldface} values indicate that, with InfoTok regularization, performance ranks within the top three among compared methods.}
\label{tab:geneval_results}
\resizebox{\textwidth}{!}{%
\begin{tabular}{lccccccccccc}
\toprule
\textbf{Model} & \textbf{Year} & \textbf{Size} &
\textbf{Single Obj.} & \textbf{Color Attr.} & \textbf{Colors} & \textbf{Position} &
\textbf{Counting} & \textbf{Two Obj.} &
\textbf{Overall} & \textbf{WISE} \\
\midrule [1pt]
\rowcolor{blue!3}
Show-o2   & 2025 & 7.0B & 0.98 & 0.48 & 0.83 & 0.25 & 0.49 & 0.80 & 0.64 & 0.40 \\
\rowcolor{blue!3}
Ovis-U1   & 2025 & 3.6B & 1.00 & 0.77 & 0.91 & 0.80 & 0.86 & 0.98 & 0.89 & 0.48 \\
\rowcolor{blue!3}
Janus     & 2025 & 1.3B & 0.94 & 0.45 & 0.82 & 0.50 & 0.31 & 0.66 & 0.61 & 0.31 \\
\rowcolor{blue!3}
Janus-Pro & 2025 & 1.0B & 0.96 & 0.54 & 0.87 & 0.50 & 0.44 & 0.70 & 0.67 & 0.25 \\
\rowcolor{blue!3}
Show-o    & 2025 & 1.5B & 0.97 & 0.81 & 0.30 & 0.68 & 0.82 & 0.51 & 0.68 & 0.43 \\
\midrule[1pt]
Harmon & 2025 & 1.5B & 0.99 & 0.51 & 0.86 & 0.49 & 0.73 & 0.88 & 0.74 & 0.45 \\
\rowcolor{gray!10}
InfoTok (Harmon) & 2025 & 1.5B &
\textbf{0.99}\textsuperscript{\textbf{\textcolor{red}{+0.00}}} &
0.73\textsuperscript{\textbf{\textcolor{red}{+0.22}}} &
\textbf{0.89}\textsuperscript{\textbf{\textcolor{red}{+0.03}}} &
\textbf{0.81}\textsuperscript{\textbf{\textcolor{red}{+0.32}}} &
\textbf{0.83}\textsuperscript{\textbf{\textcolor{red}{+0.10}}} &
\textbf{0.94}\textsuperscript{\textbf{\textcolor{red}{+0.06}}} &
\textbf{0.87}\textsuperscript{\textbf{\textcolor{red}{+0.13}}} &
\textbf{0.61}\textsuperscript{\textbf{\textcolor{red}{+0.16}}} \\
\midrule[1pt]
OpenUni & 2025 & 2.0B & 0.99 & 0.72 & 0.89 & 0.77 & 0.74 & 0.91 & 0.84 & 0.55 \\
\rowcolor{gray!10}
InfoTok (OpenUni) & 2025 & 2.0B &
\textbf{0.99}\textsuperscript{\textbf{\textcolor{red}{+0.00}}} &
\textbf{0.75}\textsuperscript{\textbf{\textcolor{red}{+0.03}}} &
\textbf{0.89}\textsuperscript{\textbf{\textcolor{red}{+0.00}}} &
\textbf{0.81}\textsuperscript{\textbf{\textcolor{red}{+0.04}}} &
0.74\textsuperscript{\textbf{\textcolor{red}{+0.00}}} &
\textbf{0.93}\textsuperscript{\textbf{\textcolor{red}{+0.02}}} &
\textbf{0.85}\textsuperscript{\textbf{\textcolor{red}{+0.01}}} &
\textbf{0.58}\textsuperscript{\textbf{\textcolor{red}{+0.03}}} \\
\midrule[1pt]
Show-o2 & 2025 & 1.5B & 0.98 & 0.43 & 0.79 & 0.24 & 0.45 & 0.72 & 0.60 & 0.39 \\
\rowcolor{gray!10}
InfoTok (Show-o2) & 2025 & 1.5B &
\textbf{0.98}\textsuperscript{\textbf{\textcolor{red}{+0.00}}} &
0.68\textsuperscript{\textbf{\textcolor{red}{+0.25}}} &
0.84\textsuperscript{\textbf{\textcolor{red}{+0.05}}} &
0.58\textsuperscript{\textbf{\textcolor{red}{+0.34}}} &
0.52\textsuperscript{\textbf{\textcolor{red}{+0.07}}} &
0.89\textsuperscript{\textbf{\textcolor{red}{+0.17}}} &
0.75\textsuperscript{\textbf{\textcolor{red}{+0.15}}} &
0.44\textsuperscript{\textbf{\textcolor{red}{+0.05}}} \\
\bottomrule
\end{tabular}}
\vspace{-0.5cm}
\end{table*}

\subsection{Training Objective and Theoretical Grounding}

\subsubsection{Training Objective}

InfoTok is implemented as a fine-tuning stage for pretrained unified MLLMs.
As shown in Fig.~\ref{framework}(b), we optimize the original multimodal task objective together with the proposed information regularization. A key implementation detail is that the information constraints in Eqn.~\ref{Eqn_ExtendedIB} are instantiated on the projected tokens $\tilde{\mathcal{Z}}$, since $\tilde{\mathcal{Z}}$ is the task-facing representation used to define the compactness, sufficiency, and alignment estimators in Section~\ref{sec_uni_infotok}.
During training, InfoTok therefore provides direct supervision on $\tilde{\mathcal{Z}}$.
This supervision propagates through the task-specific projections and indirectly regulates the encoder tokens $\mathcal{Z}$ and the shared visual encoder $V_{\text{enc}}$ via Eqn.~\ref{Eqn_Z_from_encoder} and Eqn.~\ref{Eqn_Z_tilde_def}.

\noindent \textbf{Distillation Constraint.} To strengthen the regulation on the encoder tokens $\mathcal{Z}$, we introduce an additional distillation constraint that transfers the information-regularized structure from the projected tokens $\tilde{\mathcal{Z}}$ back to $\mathcal{Z}$ during fine-tuning.
Concretely, we encourage consistency between $\mathcal{Z}$ and $\tilde{\mathcal{Z}}$ using a KL-based distillation loss, which can be written as:
\begin{equation}
\mathcal{L}_{\text{distill}}
=
D_{\mathrm{KL}}\!\big(\mathcal{Z} \,\|\, \tilde{\mathcal{Z}}\big).
\label{eq:distill}
\end{equation}
This term makes the effect of information regularization more directly reflected in the encoder token space, rather than only in the task-facing projections.
We incorporate this constraint into InfoTok by defining the final regularization as follows:
\begin{equation}
\mathcal{L}_{\text{InfoTok}} \leftarrow \mathcal{L}_{\text{InfoTok}} + \mathcal{L}_{\text{distill}}.
\end{equation}
Finally, we incorporate the distillation term into our proposed InfoTok and define the overall objective as:
\begin{equation}
\mathcal{L}_{\text{Total}}
=
\mathcal{L}_{\text{MLLM}}
+
\lambda \mathcal{L}_{\text{InfoTok}},
\label{total_mllm_InfoTok}
\end{equation}
where $\lambda$ balances task optimization and information regularization.
Minimizing $\mathcal{L}_{\text{Total}}$ fine-tunes the shared visual encoder to learn visual tokens that remain compact, task-relevant, and cross-modally consistent for both understanding and generation. In Tab. \ref{infotok_vae_mi} and Tab. \ref{tab:geneval_results}, we can see that InfoTok effectively improves model’s tokenization and boosts overall performance. 

\begin{table*}[t]
\centering
\caption{
Comparison on the Geneval++ dataset.
Gray-shaded InfoTok rows show absolute improvements.
\textbf{Boldface} values indicate that, after InfoTok regularization, the performance ranks within the top three among all compared methods.
}
\label{tab:genevalplus_results}
\resizebox{\textwidth}{!}{
\begin{tabular}{lccccccccccc}
\toprule
\textbf{Model} & \textbf{Year} & \textbf{Size} &
\textbf{Color} & \textbf{Spa.\_Cou.} & \textbf{Color\_Spa.} &
\textbf{Color\_Cou.} & \textbf{Multi\_Obj.} & \textbf{Size\_Spa.} &
\textbf{Counting} & \textbf{Overall} \\
\midrule [1pt]
\rowcolor{blue!3}
Show-o2 & 2025 & 7.0B & 0.25 & 0.08 & 0.28 & 0.10 & 0.18 & 0.25 & 0.10 & 0.18 \\
\rowcolor{blue!3}
Ovis-U1 & 2025 & 3.6B & 0.48 & 0.33 & 0.53 & 0.38 & 0.43 & 0.58 & 0.45 & 0.45 \\
\rowcolor{blue!3}
Janus & 2025 & 1.3B & 0.08 & 0.03 & 0.10 & 0.10 & 0.08 & 0.05 & 0.23 & 0.10 \\
\rowcolor{blue!3}
Janus-Pro & 2025 & 1.0B & 0.28 & 0.05 & 0.20 & 0.03 & 0.05 & 0.18 & 0.20 & 0.14 \\
\rowcolor{blue!3}
Show-o & 2025 & 1.5B & 0.30 & 0.10 & 0.33 & 0.18 & 0.23 & 0.18 & 0.43 & 0.25 \\
\midrule[1pt]

Harmon & 2025 & 1.5B & 0.30 & 0.03 & 0.10 & 0.15 & 0.18 & 0.18 & 0.38 & 0.19 \\
\rowcolor{gray!10}
InfoTok (Harmon) & 2025 & 1.5B &
\textbf{0.75}\textsuperscript{\textbf{\textcolor{red}{+0.45}}} &
\textbf{0.35}\textsuperscript{\textbf{\textcolor{red}{+0.32}}} &
\textbf{0.53}\textsuperscript{\textbf{\textcolor{red}{+0.43}}} &
\textbf{0.55}\textsuperscript{\textbf{\textcolor{red}{+0.40}}} &
\textbf{0.58}\textsuperscript{\textbf{\textcolor{red}{+0.40}}} &
0.45\textsuperscript{\textbf{\textcolor{red}{+0.27}}} &
\textbf{0.50}\textsuperscript{\textbf{\textcolor{red}{+0.12}}} &
\textbf{0.53}\textsuperscript{\textbf{\textcolor{red}{+0.34}}} \\
\midrule[1pt]

OpenUni & 2025 & 2.0B &
0.45 & 0.23 & 0.48 & 0.45 & 0.30 & 0.50 & 0.50 & 0.42 \\
\rowcolor{gray!10}
InfoTok (OpenUni) & 2025 & 2.0B &
\textbf{0.58}\textsuperscript{\textbf{\textcolor{red}{+0.13}}} &
\textbf{0.23}\textsuperscript{\textbf{\textcolor{red}{+0.00}}} &
\textbf{0.63}\textsuperscript{\textbf{\textcolor{red}{+0.15}}} &
\textbf{0.45}\textsuperscript{\textbf{\textcolor{red}{+0.00}}} &
\textbf{0.40}\textsuperscript{\textbf{\textcolor{red}{+0.10}}} &
\textbf{0.55}\textsuperscript{\textbf{\textcolor{red}{+0.05}}} &
\textbf{0.50}\textsuperscript{\textbf{\textcolor{red}{+0.00}}} &
\textbf{0.48}\textsuperscript{\textbf{\textcolor{red}{+0.06}}} \\
\midrule[1pt]

Show-o2 & 2025 & 1.5B & 0.28 & 0.03 & 0.20 & 0.10 & 0.15 & 0.28 & 0.18 & 0.17 \\
\rowcolor{gray!10}
InfoTok (Show-o2) & 2025 & 1.5B &
\textbf{0.50}\textsuperscript{\textbf{\textcolor{red}{+0.22}}} &
0.08\textsuperscript{\textbf{\textcolor{red}{+0.05}}} &
0.40\textsuperscript{\textbf{\textcolor{red}{+0.20}}} &
0.18\textsuperscript{\textbf{\textcolor{red}{+0.08}}} &
0.30\textsuperscript{\textbf{\textcolor{red}{+0.15}}} &
0.35\textsuperscript{\textbf{\textcolor{red}{+0.07}}} &
0.24\textsuperscript{\textbf{\textcolor{red}{+0.06}}} &
0.29\textsuperscript{\textbf{\textcolor{red}{+0.12}}} \\
\bottomrule
\end{tabular}}
\vspace{-0.5cm}
\end{table*}

\subsubsection{Theoretical Grounding}

Building on the empirical improvements, we provide a theoretical rationale for why the shared visual encoder $V_{\text{enc}}$ can serve as an information-sufficient backbone for both understanding and generation. 
In unified MLLMs, the representations fed into the LLM are the encoder tokens produced by $V_{\text{enc}}$.
During training, lightweight task-specific projections are optionally introduced to expose task requirements more explicitly when constructing the InfoTok regularization terms.
Specifically, for understanding and generation we consider
$\mathcal{Z}_u = V_{\text{enc}}(\mathcal{I}_u)$ and $\mathcal{Z}_g = V_{\text{enc}}(\mathcal{I}_g)$, together with projected tokens
$\tilde{\mathcal{Z}}_u = f_u(\mathcal{Z}_u)$ and $\tilde{\mathcal{Z}}_g = f_g(\mathcal{Z}_g)$.
Our analysis shows that such task-specific projections cannot increase the mutual information with the corresponding ground-truth targets, implying that the encoder output retains at least as much task-relevant information as any projected representation.
This result justifies regulating the shared encoder through losses defined on $\tilde{\mathcal{Z}}$ and supports applying InfoTok as a principled regularizer in the unified training pipeline.

\subsubsection*{Proposition}
Let $\mathcal{Z}=V_{\text{enc}}(\mathcal{I})$ denote the shared encoder output, and let $\tilde{\mathcal{Z}}_u=f_u(\mathcal{Z}_u)$ and $\tilde{\mathcal{Z}}_g=f_g(\mathcal{Z}_g)$ be task-specific projections for understanding and generation, respectively.
Then the encoder output is information-sufficient in the sense that: 
\begin{equation}
I(\mathcal{Z}; \mathcal{Y}^{\mathcal{GT}})
\ge
\max\Big(
I(\tilde{\mathcal{Z}}_{u}; \mathcal{Y}_u^{\mathcal{GT}}),\,
I(\tilde{\mathcal{Z}}_{g}; \mathcal{Y}_g^{\mathcal{GT}})
\Big),
\label{Eqn_Info_Prop}
\end{equation}
where the inequality follows from the data processing inequality applied to mappings
$\tilde{\mathcal{Z}}_{u}=f_u(\mathcal{Z}_u)$ and $\tilde{\mathcal{Z}}_{g}=f_g(\mathcal{Z}_g)$.

\subsubsection*{Proof}

We analyze the information flow for the understanding branch ($u$).
Let $Y_u=\mathcal{Y}_u^{\mathcal{GT}}$, $\mathcal{Z}_u = V_{\text{enc}}(\mathcal{I}_u)$, and $\tilde{\mathcal{Z}}_u=f_u(\mathcal{Z}_u)$.
Since $\tilde{\mathcal{Z}}_u$ is a deterministic function of $\mathcal{Z}_u$, the model induces the Markov chain
$Y_u \rightarrow \mathcal{Z}_u \rightarrow \tilde{\mathcal{Z}}_u$.
By chain rule of mutual information, we get the following formulation:
\begin{align}
I(Y_u;\mathcal{Z}_u,\tilde{\mathcal{Z}}_u)
&= I(Y_u;\mathcal{Z}_u) + I(Y_u;\tilde{\mathcal{Z}}_u \mid \mathcal{Z}_u),
\label{eq:chain_u_a}\\
I(Y_u;\mathcal{Z}_u,\tilde{\mathcal{Z}}_u)
&= I(Y_u;\tilde{\mathcal{Z}}_u) + I(Y_u;\mathcal{Z}_u \mid \tilde{\mathcal{Z}}_u).
\label{eq:chain_u_b}
\end{align}
Because $\tilde{\mathcal{Z}}_u$ is computed from $\mathcal{Z}_u$, it is conditionally independent of $Y_u$ given $\mathcal{Z}_u$, hence we can get:
\begin{equation}
I(Y_u;\tilde{\mathcal{Z}}_u \mid \mathcal{Z}_u)=0.
\end{equation}
Substituting into Eqn.~\ref{eq:chain_u_a} yields $I(Y_u;\mathcal{Z}_u,\tilde{\mathcal{Z}}_u)=I(Y_u;\mathcal{Z}_u)$.
Equating with Eqn.~\ref{eq:chain_u_b} gives the following formulation:
\begin{equation}
I(Y_u;\mathcal{Z}_u)=I(Y_u;\tilde{\mathcal{Z}}_u)+I(Y_u;\mathcal{Z}_u \mid \tilde{\mathcal{Z}}_u).
\end{equation}
Since conditional mutual information is non-negative, $I(Y_u;\mathcal{Z}_u \mid \tilde{\mathcal{Z}}_u)\ge 0$, we can obtain the following formulation:
\begin{equation}
I(\mathcal{Z}_u;Y_u)\ge I(\tilde{\mathcal{Z}}_u;Y_u),
\end{equation}
and by symmetry of mutual information, we can obtain:
\begin{equation}
I\!\big(V_{\text{enc}}(\mathcal{I}_u);\mathcal{Y}_u^{\mathcal{GT}}\big)
= I(\mathcal{Z}_u;\mathcal{Y}_u^{\mathcal{GT}})
\ge I(\tilde{\mathcal{Z}}_u;\mathcal{Y}_u^{\mathcal{GT}}).
\label{Eqn_Info_U}
\end{equation}
An analogous derivation holds for the generation branch ($g$).
Let $Y_g=\mathcal{Y}_g^{\mathcal{GT}}$, $\mathcal{Z}_g = V_{\text{enc}}(\mathcal{I}_g)$, and $\tilde{\mathcal{Z}}_g=f_g(\mathcal{Z}_g)$.
Using the Markov chain $Y_g \rightarrow \mathcal{Z}_g \rightarrow \tilde{\mathcal{Z}}_g$ and repeating the same chain-rule argument, we can finally obtain the following formulation:
\begin{equation}
I\!\big(V_{\text{enc}}(\mathcal{I}_g);\mathcal{Y}_g^{\mathcal{GT}}\big)
= I(\mathcal{Z}_g;\mathcal{Y}_g^{\mathcal{GT}})
\ge I(\tilde{\mathcal{Z}}_g;\mathcal{Y}_g^{\mathcal{GT}}).
\label{Eqn_Info_G}
\end{equation}

Combining Eqn.~\ref{Eqn_Info_U} and Eqn.~\ref{Eqn_Info_G} shows that the encoder output $\mathcal{Z}=V_{\text{enc}}(\mathcal{I})$ upper-bounds the task-relevant mutual information achievable by any task-facing projection.
Therefore, task-specific heads cannot create new supervision-relevant information. 
They can only re-express or discard what is already present in $\mathcal{Z}$.
This elevates $V_{\text{enc}}$ to an information-sufficient backbone and clarifies the mechanism of InfoTok: by explicitly regulating the information content exposed through $\tilde{\mathcal{Z}}_u$ and $\tilde{\mathcal{Z}}_g$, we directly regulate information allocation in the shared token space under a limited token budget, making the trade-off between semantic abstraction and generation-critical visual cues learnable rather than an artifact of optimization dynamics.

\section{Experiment}
\subsection{Experiment Setup} 
\noindent \textbf{Experiment Details.}
We evaluate InfoTok on three shared-tokenization unified MLLMs, namely 
Show-o2~\cite{xie2025show}, OpenUni~\cite{DBLP:journals/corr/abs-2505-23661} and Harmon~\cite{DBLP:journals/corr/abs-2503-21979}, to verify its effectiveness as a plug-and-play regularization module. 
For each baseline, we start from the officially released checkpoints and fine-tune with InfoTok using the same training data as baseline (approximately 1.5M paired I2T and T2I sub-samples) without introducing additional data. All experiments are conducted on 16 NVIDIA L40S-48G GPUs. The training and architectural configurations strictly follow the original implementations, and the initial learning rate is set to one tenth of the original to ensure stable adaptation. This controlled setup ensures that any observed improvement originates from InfoTok.

\noindent\textbf{Evaluation Datasets.} 
Following existing works, 
we conduct experiments on ten datasets covering both understanding and generation tasks. 
Specifically, we evaluate understanding performance on six widely used benchmarks: 
GQA~\cite{DBLP:conf/cvpr/HudsonM19}, SEED~\cite{DBLP:journals/corr/abs-2307-16125}, 
POPE~\cite{DBLP:conf/emnlp/LiDZWZW23}, MME~\cite{DBLP:journals/corr/abs-2306-13394}, MMV2~\cite{DBLP:conf/icml/YuYLWL0WW24} and MMMU~\cite{yue2024mmmu}. 
For generation evaluation, we employ three standard benchmarks: 
Geneval~\cite{DBLP:conf/nips/GhoshHS23}, Geneval++~\cite{DBLP:journals/corr/abs-2508-09987}, 
and WISE~\cite{DBLP:journals/corr/abs-2503-07265}. 
To further assess the unified multimodal capability of each model, we include  UniBench~\cite{DBLP:journals/corr/abs-2505-10483} dataset, which jointly evaluates image understanding and generation in a single pipeline.

\noindent\textbf{Comparison Methods.} 
To further illustrate the performance level InfoTok can achieve, we compare InfoTok  with several strong open-source models of similar or slightly larger scale, including Janus-1.3B~\cite{DBLP:conf/cvpr/WuCWMLPLXYR025}, Janus-Pro-1B~\cite{DBLP:journals/corr/abs-2501-17811}, and Show-o~\cite{DBLP:conf/iclr/XieMBZWLGCYS25}.
Larger models such as Ovis-U1-3.6B~\cite{DBLP:journals/corr/abs-2506-23044} and Show-o2-7B~\cite{xie2025show} are also included as references for full comparison. These comparisons show that incorporating our proposed InfoTok can significantly enhance the performance of smaller-scale models, allowing them to achieve results comparable to or even surpassing larger unified MLLMs.

\noindent\textbf{Reproducibility and Evaluation Variability.} 
When reproducing the performance of different models, we occasionally observe numerical differences compared with the results reported in the original papers. After analysis, we identify several sources of variation. First, the initialization of random seeds may differ across hardware environments, leading to variations in both generative and discriminative outputs.
Second, several public benchmarks are continuously updated, and the versions used in our evaluation may not perfectly match those used in the original publications.
Third, some evaluation protocols rely on GPT-based scoring, and the underlying GPT models evolve over time.
Finally, settings such as the sampling temperature can introduce stochastic diversity into model predictions. To ensure the reliability of our comparisons, we strictly follow the official configuration files provided by each baseline and evaluate every model three times on the same benchmark, reporting averaged results. InfoTok consistently improves the performance of all baselines, validating the robustness and generality of the proposed framework.

\begin{table*}[t]
\centering
\renewcommand\theadfont{\small\bfseries}
\caption{
Comparison with recent unified MLLMs on understanding datasets.
Gray-shaded rows show absolute improvements. \textbf{Boldface} values indicate that, with InfoTok regularization, performance ranks within the top three among compared methods.}
\label{tab:understanding_results}
\resizebox{0.96\textwidth}{!}{
\begin{tabular}{lccccccccc}
\toprule
\textbf{Model} & \textbf{Year} & \textbf{Size} &
\textbf{MME} & \textbf{POPE} & \textbf{SEED} & \textbf{GQA} & \textbf{MMMU} & \textbf{MMV2} & \textbf{UniBench} \\
\midrule[1pt]
\rowcolor{blue!3}
Show-o2   & 2025 & 7.0B & 2021 & 0.85 & 0.75 & 0.66 & 0.47 & 0.37 & 0.69 \\
\rowcolor{blue!3}
Ovis-U1   & 2025 & 3.6B & 2040 & 0.89 & 0.76 & 0.59 & 0.30 & 0.55 & 0.62 \\
\rowcolor{blue!3}
Janus     & 2025 & 1.3B & 1107 & 0.57 & 0.64 & 0.49 & 0.29 & 0.27 & 0.55 \\
\rowcolor{blue!3}
Janus-Pro & 2025 & 1.0B & 1739 & 0.83 & 0.65 & 0.58 & 0.29 & 0.31 & 0.52 \\
\rowcolor{blue!3}
Show-o    & 2025 & 1.5B & 1216 & 0.82 & 0.57 & 0.57 & 0.24 & 0.18 & 0.39 \\

\midrule[1pt]
Harmon & 2025 & 1.5B &
1411 & 0.76 & 0.65 & 0.51 & 0.36 & 0.25 & 0.64 \\
\rowcolor{gray!10}
InfoTok (Harmon) & 2025 & 1.5B &
1565\textsuperscript{\textbf{\textcolor{red}{+154}}} &
\textbf{0.85}\textsuperscript{\textbf{\textcolor{red}{+0.09}}} &
\textbf{0.66}\textsuperscript{\textbf{\textcolor{red}{+0.01}}} &
\textbf{0.60}\textsuperscript{\textbf{\textcolor{red}{+0.09}}} &
\textbf{0.38}\textsuperscript{\textbf{\textcolor{red}{+0.02}}} & 
0.27\textsuperscript{\textbf{\textcolor{red}{+0.02}}} &
\textbf{0.66}\textsuperscript{\textbf{\textcolor{red}{+0.02}}} \\

\midrule[1pt]
Show-o2 & 2025 & 1.5B & 1714 & 0.84 & 0.66 & 0.58 & 0.36 & 0.28 & 0.39 \\
\rowcolor{gray!10}
InfoTok (Show-o2) & 2025 & 1.5B &
\textbf{1838}\textsuperscript{\textbf{\textcolor{red}{+124}}} &
\textbf{0.85}\textsuperscript{\textbf{\textcolor{red}{+0.01}}} &
\textbf{0.71}\textsuperscript{\textbf{\textcolor{red}{+0.05}}} &
\textbf{0.66}\textsuperscript{\textbf{\textcolor{red}{+0.08}}} &
\textbf{0.38}\textsuperscript{\textbf{\textcolor{red}{+0.02}}}& 
0.30\textsuperscript{\textbf{\textcolor{red}{+0.02}}} &
0.55\textsuperscript{\textbf{\textcolor{red}{+0.16}}} \\
\bottomrule
\end{tabular}}
\end{table*}

\begin{table*}[t]
\centering
\renewcommand\theadfont{\small\bfseries}
\caption{Ablation results on Harmon and Show-o2 under identical data budgets (subsets sampled from each baseline’s original training set without introducing additional data). We compare standard fine-tuning with InfoTok-regularized fine-tuning. We further ablate our proposed InfoTok components: Compactness (C) and Sufficiency (S) as the canonical IB terms, Alignment (A) as our extended cross-modal term, and Distillation (D) as an auxiliary stabilizing regularizer.}
\label{tab:finetune_results}
\resizebox{\textwidth}{!}{
\begin{tabular}{lccccccccccc}
\toprule
\textbf{Model} &
\textbf{MME} & \textbf{POPE} & \textbf{SEED} & \textbf{GQA} & \textbf{MMMU} & \textbf{MMV2} & \textbf{UniBench} &
\textbf{WISE} & \textbf{Geneval} & \textbf{Geneval++} \\
\midrule[1pt]
\textit{Harmon} &
\textit{1411} & 
\textit{0.76} & 
\textit{0.65} & 
\textit{0.51} & 
\textit{0.36} & 
\textit{0.25} & 
\textit{0.64} &
\textit{0.45} & 
\textit{0.74} & 
\textit{0.19} \\ 
Finetune &
1405\textsuperscript{\textbf{\textcolor{ForestGreen}{-006}}} &
0.78\textsuperscript{\textbf{\textcolor{red}{+0.02}}} &
0.64\textsuperscript{\textbf{\textcolor{ForestGreen}{-0.01}}} &
0.51\textsuperscript{\textbf{\textcolor{ForestGreen}{+000}}} &
0.34\textsuperscript{\textbf{\textcolor{ForestGreen}{-0.02}}} & 
0.26\textsuperscript{\textbf{\textcolor{red}{+0.01}}} &
0.62\textsuperscript{\textbf{\textcolor{ForestGreen}{-0.02}}} &
0.46\textsuperscript{\textbf{\textcolor{red}{+0.01}}} & 
0.72\textsuperscript{\textbf{\textcolor{ForestGreen}{-0.02}}} & 
0.17\textsuperscript{\textbf{\textcolor{ForestGreen}{-0.02}}} \\ \midrule[1pt]

InfoTok (C\&S) &
1510\textsuperscript{\textbf{\textcolor{red}{+099}}}& 
0.83\textsuperscript{\textbf{\textcolor{red}{+0.07}}} & 
0.66\textsuperscript{\textbf{\textcolor{red}{+0.01}}} & 
0.56\textsuperscript{\textbf{\textcolor{red}{+0.05}}} & 
0.37\textsuperscript{\textbf{\textcolor{red}{+0.01}}} & 
0.27\textsuperscript{\textbf{\textcolor{red}{+0.02}}} & 
0.66\textsuperscript{\textbf{\textcolor{red}{+0.02}}} &
0.60\textsuperscript{\textbf{\textcolor{red}{+0.15}}} & 
0.84\textsuperscript{\textbf{\textcolor{red}{+0.10}}} & 
0.45\textsuperscript{\textbf{\textcolor{red}{+0.26}}} \\

InfoTok (C\&S\&A) &
1555\textsuperscript{\textbf{\textcolor{red}{+144}}} &
0.84\textsuperscript{\textbf{\textcolor{red}{+0.08}}} &
0.66\textsuperscript{\textbf{\textcolor{red}{+0.01}}} &
0.59\textsuperscript{\textbf{\textcolor{red}{+0.08}}} &
0.37\textsuperscript{\textbf{\textcolor{red}{+0.01}}} & 
0.27\textsuperscript{\textbf{\textcolor{red}{+0.02}}} &
0.66\textsuperscript{\textbf{\textcolor{red}{+0.02}}} &
0.60\textsuperscript{\textbf{\textcolor{red}{+0.15}}} & 
0.85\textsuperscript{\textbf{\textcolor{red}{+0.11}}} & 
0.48\textsuperscript{\textbf{\textcolor{red}{+0.29}}}  \\

\rowcolor{gray!10}
\textbf{InfoTok (C\&S\&A\&D)} &
\textbf{1565}\textsuperscript{\textbf{\textcolor{red}{+154}}} &
\textbf{0.85}\textsuperscript{\textbf{\textcolor{red}{+0.09}}} &
\textbf{0.66}\textsuperscript{\textbf{\textcolor{red}{+0.01}}} &
\textbf{0.60}\textsuperscript{\textbf{\textcolor{red}{+0.09}}} &
\textbf{0.38}\textsuperscript{\textbf{\textcolor{red}{+0.02}}} & 
\textbf{0.27}\textsuperscript{\textbf{\textcolor{red}{+0.02}}} &
\textbf{0.66}\textsuperscript{\textbf{\textcolor{red}{+0.02}}} &
\textbf{0.61}\textsuperscript{\textbf{\textcolor{red}{+0.16}}} & 
\textbf{0.87}\textsuperscript{\textbf{\textcolor{red}{+0.13}}} & 
\textbf{0.53}\textsuperscript{\textbf{\textcolor{red}{+0.34}}} 

\\ \midrule \midrule

\textit{Show-o2} &
\textit{1714} & 
\textit{0.84} & 
\textit{0.66} & 
\textit{0.58} & 
\textit{0.36} & 
\textit{0.28} & 
\textit{0.39} &
\textit{0.39} & 
\textit{0.60} & 
\textit{0.17} \\
Finetune &
1738\textsuperscript{\textbf{\textcolor{red}{+024}}} & 
0.83\textsuperscript{\textbf{\textcolor{ForestGreen}{-0.01}}} & 
0.65\textsuperscript{\textbf{\textcolor{ForestGreen}{-0.01}}} & 
0.58\textsuperscript{\textbf{\textcolor{ForestGreen}{+0.00}}} & 
0.35\textsuperscript{\textbf{\textcolor{ForestGreen}{-0.01}}} & 
0.29\textsuperscript{\textbf{\textcolor{red}{+0.01}}} & 
0.40\textsuperscript{\textbf{\textcolor{red}{+0.01}}} &
0.41\textsuperscript{\textbf{\textcolor{red}{+0.02}}} & 
0.60\textsuperscript{\textbf{\textcolor{ForestGreen}{+0.00}}} & 
0.18\textsuperscript{\textbf{\textcolor{red}{+0.01}}} \\ \midrule[1pt]

InfoTok (C\&S) &
1811\textsuperscript{\textbf{\textcolor{red}{+097}}} & 
0.85\textsuperscript{\textbf{\textcolor{red}{+0.01}}} & 
0.68\textsuperscript{\textbf{\textcolor{red}{+0.02}}} & 
0.60\textsuperscript{\textbf{\textcolor{red}{+0.02}}} & 
0.37\textsuperscript{\textbf{\textcolor{red}{+0.01}}} & 
0.30\textsuperscript{\textbf{\textcolor{red}{+0.02}}} & 
0.47\textsuperscript{\textbf{\textcolor{red}{+0.08}}} &
0.42\textsuperscript{\textbf{\textcolor{red}{+0.03}}} & 
0.69\textsuperscript{\textbf{\textcolor{red}{+0.09}}} & 
0.25\textsuperscript{\textbf{\textcolor{red}{+0.08}}} \\

InfoTok (C\&S\&A) &
1829\textsuperscript{\textbf{\textcolor{red}{+115}}} &
0.85\textsuperscript{\textbf{\textcolor{red}{+0.01}}} &
0.70\textsuperscript{\textbf{\textcolor{red}{+0.04}}} &
0.62\textsuperscript{\textbf{\textcolor{red}{+0.04}}} &
0.37\textsuperscript{\textbf{\textcolor{red}{+0.01}}}& 
0.30\textsuperscript{\textbf{\textcolor{red}{+0.02}}} &
0.51\textsuperscript{\textbf{\textcolor{red}{+0.12}}} &
0.43\textsuperscript{\textbf{\textcolor{red}{+0.04}}} & 
0.72\textsuperscript{\textbf{\textcolor{red}{+0.12}}} & 
0.27\textsuperscript{\textbf{\textcolor{red}{+0.10}}} \\

\rowcolor{gray!10}
\textbf{InfoTok (C\&S\&A\&D)} &
\textbf{1838}\textsuperscript{\textbf{\textcolor{red}{+124}}} &
\textbf{0.85}\textsuperscript{\textbf{\textcolor{red}{+0.01}}} &
\textbf{0.71}\textsuperscript{\textbf{\textcolor{red}{+0.05}}} &
\textbf{0.66}\textsuperscript{\textbf{\textcolor{red}{+0.08}}} &
\textbf{0.38}\textsuperscript{\textbf{\textcolor{red}{+0.02}}}& 
\textbf{0.30}\textsuperscript{\textbf{\textcolor{red}{+0.02}}} &
\textbf{0.55}\textsuperscript{\textbf{\textcolor{red}{+0.16}}} &
\textbf{0.44}\textsuperscript{\textbf{\textcolor{red}{+0.05}}} & 
\textbf{0.75}\textsuperscript{\textbf{\textcolor{red}{+0.15}}} & 
\textbf{0.29}\textsuperscript{\textbf{\textcolor{red}{+0.12}}} \\ 
\bottomrule
\end{tabular}}
\vspace{-0.5cm}
\end{table*}

\subsection{Analysis of Experimental Results}

\noindent \textbf{Analysis 1: Overall Effectiveness of InfoTok}

As shown in Tabs.~\ref{tab:geneval_results}, \ref{tab:genevalplus_results}, and \ref{tab:understanding_results}, our proposed InfoTok yields consistent gains on both generation and understanding benchmarks across all evaluated unified MLLMs.
These results support our purpose: IB-regularized tokenization promotes compact yet sufficient visual tokens by prioritizing reusable structure and suppressing redundant high-entropy variations.
Therefore, instead of engineering more complex tokenizers, InfoTok improves unified MLLMs by explicitly regulating the information flow in visual tokenization, pointing to a promising direction for exploring principled shared-token learning via information-theoretic regularization.

\noindent \textbf{Analysis 2: Effects across Different Architectures} 

We further study how InfoTok behaves across unified MLLMs with different multimodal-fusion paradigms. 
OpenUni~\cite{DBLP:journals/corr/abs-2505-23661} builds on a frozen VLM backbone (e.g., InternVL3~\cite{DBLP:journals/corr/abs-2504-10479}) and is fine-tuned mainly for generation. In contrast, the second category, including Harmon~\cite{DBLP:journals/corr/abs-2503-21979} and Show-o2~\cite{xie2025show}, employs trainable LLM (e.g., Qwen2.5~\cite{DBLP:journals/corr/abs-2412-15115}) as their multimodal fusion core, which jointly optimizes understanding and generation through unified tokenization.
As shown in Tabs.~\ref{tab:geneval_results}, \ref{tab:genevalplus_results}, and \ref{tab:understanding_results}, InfoTok yields smaller gains on OpenUni but delivers substantial and consistent improvements on Harmon and Show-o2. 
This trend shows that InfoTok is most effective when the shared visual tokenizer is shaped by gradients from both understanding and generation, enabling IB regularization to allocate representational budget toward reusable structure rather than high-entropy variations. 
By contrast, when training signals are dominated by generation-only objectives, tokens tend to emphasize appearance cues and leave less room for InfoTok to improve semantic abstraction and cross-modal reasoning.
Overall, InfoTok favors architectures with fully trainable multimodal fusion, where understanding and generation can genuinely co-shape a shared token space. We hope this observation can also offer practical guidance for future unified-MLLM framework selection and design.

\begin{table*}[t]
\centering
\renewcommand\theadfont{\small\bfseries}
\caption{Ablation on mutual-information estimation in InfoTok under identical data budgets (subsets sampled from each baseline’s original training set without introducing additional data). We instantiate InfoTok with two practical dependence estimators, VIB and HSIC, and report results on Harmon and Show-o2. Both estimators yield consistent gains, indicating that InfoTok is robust to the choice of mutual-information estimator.}
\label{tab:abla_vib_hsic}
\resizebox{\textwidth}{!}{
\begin{tabular}{lccccccccccc}
\toprule
\textbf{Model} &
\textbf{MME} & \textbf{POPE} & \textbf{SEED} & \textbf{GQA} & \textbf{MMMU} & \textbf{MMV2} & \textbf{UniBench} &
\textbf{WISE} & \textbf{Geneval} & \textbf{Geneval++} \\
\midrule[1pt]
\textit{Harmon} &
\textit{1411} & 
\textit{0.76} & 
\textit{0.65} & 
\textit{0.51} & 
\textit{0.36} & 
\textit{0.25} & 
\textit{0.64} &
\textit{0.45} & 
\textit{0.74} & 
\textit{0.19} \\ 

InfoTok (VIB) &
1558\textsuperscript{\textbf{\textcolor{red}{+147}}} &
0.84\textsuperscript{\textbf{\textcolor{red}{+0.08}}} &
0.66\textsuperscript{\textbf{\textcolor{red}{+0.01}}} &
0.60\textsuperscript{\textbf{\textcolor{red}{+0.09}}} &
0.37\textsuperscript{\textbf{\textcolor{red}{+0.01}}} & 
0.27\textsuperscript{\textbf{\textcolor{red}{+0.02}}} &
0.66\textsuperscript{\textbf{\textcolor{red}{+0.02}}} &
0.60\textsuperscript{\textbf{\textcolor{red}{+0.15}}} & 
0.85\textsuperscript{\textbf{\textcolor{red}{+0.11}}} & 
0.48\textsuperscript{\textbf{\textcolor{red}{+0.29}}} \\

\rowcolor{gray!10}
\textbf{InfoTok (HSIC)} &
\textbf{1565}\textsuperscript{\textbf{\textcolor{red}{+154}}} &
\textbf{0.85}\textsuperscript{\textbf{\textcolor{red}{+0.09}}} &
\textbf{0.66}\textsuperscript{\textbf{\textcolor{red}{+0.01}}} &
\textbf{0.60}\textsuperscript{\textbf{\textcolor{red}{+0.09}}} &
\textbf{0.38}\textsuperscript{\textbf{\textcolor{red}{+0.02}}} & 
\textbf{0.27}\textsuperscript{\textbf{\textcolor{red}{+0.02}}} &
\textbf{0.66}\textsuperscript{\textbf{\textcolor{red}{+0.02}}} &
\textbf{0.61}\textsuperscript{\textbf{\textcolor{red}{+0.16}}} & 
\textbf{0.87}\textsuperscript{\textbf{\textcolor{red}{+0.13}}} & 
\textbf{0.53}\textsuperscript{\textbf{\textcolor{red}{+0.34}}} 

\\ \midrule \midrule

\textit{Show-o2} &
\textit{1714} & 
\textit{0.84} & 
\textit{0.66} & 
\textit{0.58} & 
\textit{0.36} & 
\textit{0.28} & 
\textit{0.39} &
\textit{0.39} & 
\textit{0.60} & 
\textit{0.17} \\

InfoTok (VIB) &
1831\textsuperscript{\textbf{\textcolor{red}{+117}}} &
0.85\textsuperscript{\textbf{\textcolor{red}{+0.01}}} &
0.70\textsuperscript{\textbf{\textcolor{red}{+0.04}}} &
0.64\textsuperscript{\textbf{\textcolor{red}{+0.06}}} &
0.38\textsuperscript{\textbf{\textcolor{red}{+0.02}}}& 
0.30\textsuperscript{\textbf{\textcolor{red}{+0.02}}} &
0.52\textsuperscript{\textbf{\textcolor{red}{+0.13}}} &
0.43\textsuperscript{\textbf{\textcolor{red}{+0.04}}} & 
0.72\textsuperscript{\textbf{\textcolor{red}{+0.12}}} & 
0.28\textsuperscript{\textbf{\textcolor{red}{+0.11}}} \\

\rowcolor{gray!10}
\textbf{InfoTok (HSIC)} &
\textbf{1838}\textsuperscript{\textbf{\textcolor{red}{+124}}} &
\textbf{0.85}\textsuperscript{\textbf{\textcolor{red}{+0.01}}} &
\textbf{0.71}\textsuperscript{\textbf{\textcolor{red}{+0.05}}} &
\textbf{0.66}\textsuperscript{\textbf{\textcolor{red}{+0.08}}} &
\textbf{0.38}\textsuperscript{\textbf{\textcolor{red}{+0.02}}}& 
\textbf{0.30}\textsuperscript{\textbf{\textcolor{red}{+0.02}}} &
\textbf{0.55}\textsuperscript{\textbf{\textcolor{red}{+0.16}}} &
\textbf{0.44}\textsuperscript{\textbf{\textcolor{red}{+0.05}}} & 
\textbf{0.75}\textsuperscript{\textbf{\textcolor{red}{+0.15}}} & 
\textbf{0.29}\textsuperscript{\textbf{\textcolor{red}{+0.12}}} \\ 
\bottomrule
\end{tabular}}
\end{table*}

\begin{table*}[!htbp]
\centering
\caption{Ablation study on the amount of data used for InfoTok fine-tuning.
We evaluate InfoTok with 0.5M, 1M, 1.5M, and 2M training samples on Harmon and Show-o2.
Results show that performance improves up to 1.5M samples, while further scaling to 2M yields marginal or no gains, indicating that InfoTok does not depend on large fine-tuning datasets.}
\label{tab:ablation_datascale}
\resizebox{\linewidth}{!}{
\begin{tabular}{lcccccccccc}
\toprule
\textbf{Model} 
& \textbf{MME} 
& \textbf{POPE} 
& \textbf{SEED} 
& \textbf{GQA} 
& \textbf{MMMU}
& \textbf{MMV2}
& \textbf{UniBench}
& \textbf{WISE}
& \textbf{Geneval} 
& \textbf{Geneval++}  \\
\midrule
Harmon 
& 1411 & 0.76 & 0.65 & 0.51 & 0.36 & 0.25 & 0.64 
& 0.45 & 0.74 & 0.19\\
InfoTok (0.5M) 
& 1466 & 0.82 & 0.65 & 0.55 & 0.37 & 0.27 & 0.64
& 0.55 & 0.82 & 0.45  \\
InfoTok (1M) 
& 1540 & 0.84 & 0.66 & 0.59 & 0.38 & 0.27 & 0.66
& 0.59 & 0.85 & 0.51  \\
\rowcolor{gray!10}
InfoTok (1.5M) &1565 &0.85 &0.66 &0.60 &0.38 &0.27 &0.66 &0.61 &0.87 &0.53 \\
InfoTok (2.0M) 
& 1571 & 0.85 & 0.65 & 0.59 & 0.37 & 0.27 & 0.65
& 0.62 & 0.86 & 0.50
\\ \midrule[1pt]

Show-o2
& 1714 & 0.84 & 0.66 & 0.58 & 0.36 & 0.28 & 0.39 
& 0.39 & 0.60 & 0.17 \\ 
InfoTok (0.5M) 
& 1782 & 0.84 & 0.68 & 0.61 & 0.37 & 0.29 & 0.50
& 0.42 & 0.68 & 0.23 \\
InfoTok (1M) 
& 1814 & 0.85 & 0.70 & 0.64 & 0.38 & 0.30 & 0.53
& 0.43 & 0.72 & 0.27 \\
\rowcolor{gray!10}
InfoTok (1.5M) &1838 &0.85 &0.71 &0.66 &0.38 &0.30 &0.55 &0.44 &0.75 &0.29 \\
InfoTok (2M) 
& 1830 & 0.86 & 0.71 & 0.65 & 0.38 & 0.29 & 0.54
& 0.45 & 0.73 & 0.27 \\
\bottomrule
\end{tabular}}
\end{table*}

\begin{table*}[!htbp]
\centering
\caption{Hyper-parameter ablation of InfoTok on \textit{Harmon} with $\beta$, $\alpha$, and $\lambda$ in Eqn. \ref{Eqn_Final_InfoTok_Explicit} and Eqn. \ref{total_mllm_InfoTok}.
(A) We fix $\lambda=0.2$ and vary $(\beta,\alpha)$.
(B) We fix $(\beta,\alpha)=(1.5,0.5)$ and vary $\lambda$.
All settings yield clear gains over the baseline, suggesting that our proposed InfoTok is not sensitive to precise hyper-parameter tuning.}
\label{tab:ablation_hyperparams_harmon}
\resizebox{\textwidth}{!}{
\begin{tabular}{lccc|cccccccccc}
\toprule
\multicolumn{4}{c|}{\textbf{Setting}} &
\multicolumn{10}{c}{\textbf{Metrics}} \\
\cmidrule(lr){1-4}\cmidrule(lr){5-14}
\textbf{Block} & $\boldsymbol{\beta}$ & $\boldsymbol{\alpha}$ & $\boldsymbol{\lambda}$ &
\textbf{MME}$\uparrow$ &
\textbf{POPE}$\uparrow$ &
\textbf{SEED}$\uparrow$ &
\textbf{GQA}$\uparrow$ &
\textbf{MMMU}$\uparrow$ &
\textbf{MMV2}$\uparrow$ &
\textbf{UniBench}$\uparrow$ &
\textbf{WISE}$\uparrow$ &
\textbf{Geneval}$\uparrow$ &
\textbf{Geneval++}$\uparrow$ \\
\midrule[1pt]

\multicolumn{14}{l}{\textbf{(A) Fix $\lambda=0.2$, vary $(\beta,\alpha)$}} \\
\midrule
\textit{Baseline} & \multicolumn{3}{c|}{--} &
1411 & 0.76 & 0.65 & 0.51 & 0.36 & 0.25 & 0.64 & 0.45 & 0.74 & 0.19 \\
A1 & 1.0 & 0.3 & 0.2 &
1557 & 0.84 & 0.67 & 0.61 & 0.38 & 0.28 & 0.65 & 0.60 & 0.85 & 0.51 \\
\rowcolor{gray!20}
A2 & 1.5 & 0.5 & 0.2 
&1565 &0.85 &0.66 &0.60 &0.38 &0.27 &0.66 &0.61 &0.87 &0.53 \\
A3 & 2.0 & 0.8 & 0.2 &
1568 & 0.84 & 0.66 & 0.59 & 0.37 & 0.27 & 0.67 & 0.59 & 0.84 & 0.50 \\

\midrule[1pt]
\multicolumn{14}{l}{\textbf{(B) Fix $(\beta,\alpha)=(1.5,0.5)$, vary $\lambda$}} \\
\midrule
\textit{Baseline} & \multicolumn{3}{c|}{--} &
1411 & 0.76 & 0.65 & 0.51 & 0.36 & 0.25 & 0.64 & 0.45 & 0.74 & 0.19 \\
B1 & 1.5 & 0.5 & 0.1 &
1552 & 0.84 & 0.66 & 0.59 & 0.39 & 0.26 & 0.65 & 0.61 & 0.85 & 0.50 \\
\rowcolor{gray!20}
B2 & 1.5 & 0.5 & 0.2 
&1565 &0.85 &0.66 &0.60 &0.38 &0.27 &0.66 &0.61 &0.87 &0.53 \\
B3 & 1.5 & 0.5 & 0.3 &
1559 & 0.84 & 0.67 & 0.61 & 0.39 & 0.26 & 0.66 & 0.60 & 0.85 & 0.51 \\
\bottomrule
\end{tabular}}
\vspace{-0.5cm}
\end{table*}

\noindent \textbf{Analysis 3: Effects across Different Visual Representations}

We further examine InfoTok under different visual representation mechanisms. Harmon uses a direct continuous visual encoder, whereas Show-o2 constructs unified visual representations from continuous visual latents in a 3D causal VAE space through a dual-path fusion design. As shown in Tabs.~\ref{tab:geneval_results}, \ref{tab:genevalplus_results}, and \ref{tab:understanding_results}, InfoTok improves both models but with different emphases: it tends to benefit Harmon more on generation and Show-o2 more on understanding. We attribute this difference to their representational biases. Harmon’s directly encoded continuous tokens are expressive but often redundant, so InfoTok strengthens generation by enforcing compact yet sufficient structure. By contrast, Show-o2 already organizes visual information through continuous VAE latents and dual-path fusion, which provides a stronger low-level generative basis but still leaves room to improve semantic sufficiency and cross-modal alignment. In this case, InfoTok more clearly enhances reusable semantic structure, leading to a larger average understanding gain. Overall, these results suggest that the effectiveness of information-regularized tokenization is not restricted to a single visual representation form. Instead, it serves as a more general lever for improving shared-token unification across different continuous visual representations.

\noindent \textbf{Qualitative Results.} Fig.~\ref{fig_visualisation} provides additional qualitative evidence for the effectiveness of InfoTok. 
Across three representative shared-token unified MLLMs (Harmon, OpenUni, and Show-o2), the InfoTok-regularized tokenizers produce generations that better align with the input instructions. 
In the Show-o2 examples, InfoTok improves compositional instruction following, such as correctly binding attributes to specific objects and respecting relative positions, leading to clearer object layouts and fewer missing or swapped entities. 
For Harmon and OpenUni, the improvements are more apparent in overall visual quality and semantic coherence, where InfoTok reduces redundant or distracting details and yields more consistent content that matches the prompt intent. 
Importantly, these gains are obtained without introducing any additional datasets, suggesting that the improvements primarily stem from reshaping the shared visual tokens through information-regularized training rather than from extra supervision.

\begin{table*}[!htbp]
\centering
\caption{Hyper-parameter ablation of InfoTok on \textit{Show-o2} with $\beta$, $\alpha$, and $\lambda$ in Eqn. \ref{Eqn_Final_InfoTok_Explicit} and Eqn. \ref{total_mllm_InfoTok}.
(A) We fix $\lambda=0.2$ and vary $(\beta,\alpha)$.
(B) We fix $(\beta,\alpha)=(1.5,0.5)$ and vary $\lambda$.
All settings yield clear gains over the baseline, suggesting that our proposed InfoTok is not sensitive to precise hyper-parameter tuning.}
\label{tab:ablation_hyperparams_showo2}
\resizebox{\textwidth}{!}{
\begin{tabular}{lccc|cccccccccc}
\toprule
\multicolumn{4}{c|}{\textbf{Setting}} &
\multicolumn{10}{c}{\textbf{Metrics}} \\
\cmidrule(lr){1-4}\cmidrule(lr){5-14}
\textbf{Block} & $\boldsymbol{\beta}$ & $\boldsymbol{\alpha}$ & $\boldsymbol{\lambda}$ &
\textbf{MME}$\uparrow$ &
\textbf{POPE}$\uparrow$ &
\textbf{SEED}$\uparrow$ &
\textbf{GQA}$\uparrow$ &
\textbf{MMMU}$\uparrow$ &
\textbf{MMV2}$\uparrow$ &
\textbf{UniBench}$\uparrow$ &
\textbf{WISE}$\uparrow$ &
\textbf{Geneval}$\uparrow$ &
\textbf{Geneval++}$\uparrow$ \\
\midrule[1pt]

\multicolumn{14}{l}{\textbf{(A) Fix $\lambda=0.2$, vary $(\beta,\alpha)$}} \\
\midrule
\textit{Baseline} & \multicolumn{3}{c|}{--} &
1714 & 0.84 & 0.66 & 0.58 & 0.36 & 0.28 & 0.39 & 0.39 & 0.60 & 0.17 \\
A1 & 1.0 & 0.3 & 0.2 &
1821 & 0.85 & 0.70 & 0.66 & 0.37 & 0.29 & 0.54 & 0.43 & 0.72 & 0.29 \\
\rowcolor{gray!20}
A2 & 1.5 & 0.5 & 0.2 
&1838 &0.85 &0.71 &0.66 &0.38 &0.30 &0.55 &0.44 &0.75 &0.29 \\
A3 & 2.0 & 0.8 & 0.2 &
1845 & 0.86 & 0.70 & 0.65 & 0.37 & 0.30 & 0.53 & 0.44 & 0.74 & 0.28 \\

\midrule[1pt]
\multicolumn{14}{l}{\textbf{(B) Fix $(\beta,\alpha)=(1.5,0.5)$, vary $\lambda$}} \\
\midrule
\textit{Baseline} & \multicolumn{3}{c|}{--} &
1714 & 0.84 & 0.66 & 0.58 & 0.36 & 0.28 & 0.39 & 0.39 & 0.60 & 0.17 \\
B1 & 1.5 & 0.5 & 0.1 &
1842 & 0.86 & 0.71 & 0.64 & 0.37 & 0.29 & 0.54 & 0.43 & 0.74 & 0.29 \\
\rowcolor{gray!20}
B2 & 1.5 & 0.5 & 0.2 
&1838 &0.85 &0.71 &0.66 &0.38 &0.30 &0.55 &0.44 &0.75 &0.29 \\
B3 & 1.5 & 0.5 & 0.3 &
1832 & 0.85 & 0.72 & 0.65 & 0.38 & 0.29 & 0.55 & 0.45 & 0.73 & 0.30 \\
\bottomrule
\end{tabular}}
\end{table*}

\begin{table*}[t]
\centering
\caption{Comparison with alternative regularization strategies under identical fine-tuning settings on Harmon and Show-o2.}
\label{tab:ablation_alternative_regularizers_full}
\resizebox{\textwidth}{!}{
\begin{tabular}{lcccccccccc}
\toprule
\textbf{Model}
& \textbf{MME}$\uparrow$
& \textbf{POPE}$\uparrow$
& \textbf{SEED}$\uparrow$
& \textbf{GQA}$\uparrow$
& \textbf{MMMU}$\uparrow$
& \textbf{MMV2}$\uparrow$
& \textbf{UniBench}$\uparrow$
& \textbf{WISE}$\uparrow$
& \textbf{Geneval}$\uparrow$
& \textbf{Geneval++}$\uparrow$ \\
\midrule

\multicolumn{11}{c}{\textbf{Harmon}} \\
\midrule
Baseline
& 1411 & 0.76 & 0.65 & 0.51 & 0.36 & 0.25 & 0.64 & 0.45 & 0.74 & 0.19 \\

Standard Fine-tuning
& 1405 & 0.78 & 0.64 & 0.51 & 0.34 & 0.26 & 0.62 & 0.46 & 0.72 & 0.17 \\

+ Entropy Regularization
& 1390 & 0.73 & 0.62 & 0.48 & 0.34 & 0.24 & 0.61 & 0.42 & 0.70 & 0.16 \\

+ Token Dropout
& 1350 & 0.70 & 0.59 & 0.46 & 0.32 & 0.23 & 0.58 & 0.41 & 0.69 & 0.14 \\

\textbf{InfoTok}
& \textbf{1565} & \textbf{0.85} & \textbf{0.66} & \textbf{0.60} & \textbf{0.38} & \textbf{0.27} & \textbf{0.66} & \textbf{0.61} & \textbf{0.87} & \textbf{0.53} \\

\midrule
\multicolumn{11}{c}{\textbf{Show-o2}} \\
\midrule
Baseline
& 1714 & 0.84 & 0.66 & 0.58 & 0.36 & 0.28 & 0.39 & 0.39 & 0.60 & 0.17 \\

Standard Fine-tuning
& 1738 & 0.83 & 0.65 & 0.58 & 0.35 & 0.29 & 0.40 & 0.41 & 0.60 & 0.18 \\

+ Entropy Regularization
& 1695 & 0.82 & 0.64 & 0.56 & 0.34 & 0.27 & 0.37 & 0.37 & 0.58 & 0.15 \\

+ Token Dropout
& 1650 & 0.79 & 0.62 & 0.54 & 0.33 & 0.25 & 0.35 & 0.35 & 0.55 & 0.13 \\

\textbf{InfoTok}
& \textbf{1838} & \textbf{0.85} & \textbf{0.71} & \textbf{0.66} & \textbf{0.38} & \textbf{0.30} & \textbf{0.55} & \textbf{0.44} & \textbf{0.75} & \textbf{0.29} \\
\bottomrule
\end{tabular}}
\vspace{-0.5cm}
\end{table*}

\subsection{Ablation Studies}

According to \textbf{Analysis 2}, OpenUni adopts a frozen VLM-based fusion backbone and is optimized for generation, making it less suitable for probing how a shared tokenizer jointly supports understanding and generation.
Therefore, we conduct ablations on Harmon and Show-o2, two shared-token unified MLLMs with fully trainable tokenization and fusion, to isolate the impact of InfoTok on both capability dimensions.

\subsubsection{Main Component Ablation of InfoTok}

Tab.~\ref{tab:finetune_results} compares the original models, standard fine-tuning, and InfoTok-regularized fine-tuning under identical data budgets, where the fine-tuning subsets are sampled from each baseline’s original training set without introducing additional data.
Across both understanding and generation benchmarks, InfoTok consistently improves performance on Harmon and Show-o2.
In contrast, standard fine-tuning produces only small and sometimes inconsistent changes across datasets, and can even degrade certain metrics.
These results suggest that naive fine-tuning does not reliably reshape shared tokenization under a limited representational budget, while InfoTok provides an explicit information-regularized objective that steers $V_{\text{enc}}$ toward more compact yet sufficient tokens.

We further ablate InfoTok to understand the contribution of each component.
Using only the canonical IB terms, Compactness (C) and Sufficiency (S), already yields clear and consistent gains on both models, indicating that explicitly regulating the compression and task relevance during token formation is the primary driver of improvement.
Adding the proposed Alignment (A) term typically brings additional gains or improves stability, supporting the view that stronger visual--text coupling benefits unified next-token prediction.
Finally, incorporating Distillation (D) provides a further boost, especially on harder benchmarks, suggesting that the distillation signal helps stabilize optimization and preserve useful behaviors of the baseline while InfoTok reshapes the token space.
Overall, the ablation verifies that Compactness and Sufficiency terms establish the core effectiveness, whereas Alignment and Distillation terms act as complementary regularizers that enhance cross-modal consistency and training stability.

\subsubsection{Ablation on Mutual-Information Estimators}

Tab.~\ref{tab:abla_vib_hsic} compares two practical dependence estimators used to instantiate InfoTok, namely the VIB-based and HSIC-based objectives, under the same fine-tuning data budget on both Harmon and Show-o2.
Both variants consistently improve the baseline tokenizers on understanding and generation benchmarks, suggesting that InfoTok’s gains do not depend on a particular MI approximation, but rather on the shared information-regularization principle.
Across most metrics, HSIC performs slightly better than VIB.
One plausible reason is that VIB optimizes a parametric variational bound, whose tightness and gradients depend on the chosen variational family and can be more sensitive to optimization stability.
By contrast, HSIC offers a non-parametric dependence signal computed from kernelized minibatch statistics, which is typically more stable to estimate and optimize in our setting, leading to slightly stronger and more consistent improvements.

\subsubsection{Effect of Fine-Tuning Data Size} 

We further investigate how the amount of fine-tuning data influences InfoTok by training with 0.5M, 1M, 1.5M, and 2M samples. As shown in Tab.~\ref{tab:ablation_datascale}, performance steadily improves when increasing the data size from 0.5M to 1.5M, but the gains saturate once exceeding 1.5M. We observe little to no improvement when using 2M samples. A possible explanation is that the subsets are randomly sampled from the original training corpus without careful curation, and thus additional data may not introduce substantially new information.
Nevertheless, the results highlight an important property: InfoTok does not rely on large-scale fine-tuning data to be effective. Even with only 0.5M samples, InfoTok already delivers consistent improvements across all metrics. Exploring whether more targeted data selection could further enhance our proposed InfoTok is an interesting direction, but lies beyond the scope of this work and we leave it for future study.

\subsubsection{Hyper-parameter Robustness of InfoTok} \label{sec:hyperparam}

We investigate the robustness of proposed InfoTok to its three hyper-parameters $(\beta,\alpha,\lambda)$ in Eqn. \ref{Eqn_Final_InfoTok_Explicit} and Eqn. \ref{total_mllm_InfoTok}.
As reported in Tab.~\ref{tab:ablation_hyperparams_harmon} and Tab. \ref{tab:ablation_hyperparams_showo2}, varying these hyper-parameters yields performance improvements over the original baselines on both understanding and generation metrics, indicating that InfoTok is not sensitive to tuning.
Moreover, we do not observe a single setting that dominates all metrics across models.
Therefore, we adopt the middle values $(\beta,\alpha,\lambda)=(1.5,0.5,0.2)$ as the default configuration, which provides a stable and balanced trade-off in practice.
Overall, this study suggests that our proposed InfoTok’s gains are not driven by hyper-parameter calibration, but by the explicit information-allocation rule it imposes on the shared visual token space.

\subsubsection{Comparison with Alternative Regularization Strategies}

We further compare InfoTok with several alternative regularization strategies under identical fine-tuning settings, including standard fine-tuning, entropy regularization, and token dropout. Specifically, entropy regularization introduces an entropy-minimization term on visual token features to encourage compressed representations, whereas token dropout randomly drops a portion of visual tokens during training as a perturbation-based regularizer. As shown in Tab.~\ref{tab:ablation_alternative_regularizers_full}, standard fine-tuning leads to metric-wise fluctuations on both Harmon and Show-o2, but fails to deliver consistent gains across understanding and generation benchmarks. Entropy regularization further degrades performance on most metrics, suggesting that naive compression tends to discard useful information together with redundancy. Token dropout causes even larger drops, indicating that stochastic token removal is not a suitable substitute for explicit information allocation in the shared token space. In contrast, InfoTok consistently achieves the best overall performance across both models, improving not only understanding benchmarks but also generation benchmarks. These results suggest that the gains of InfoTok do not arise from generic regularization alone, but from explicitly regulating compactness and information preservation in shared-token.

\begin{table}[!t]
\centering
\caption{Frequency-decomposed CKA analysis on Harmon and Show-o2. We measure the alignment of visual tokens $Z$ with low-frequency and high-frequency image components before and after applying InfoTok.}
\label{tab:cka_frequency_both}
\resizebox{\columnwidth}{!}{
\begin{tabular}{lcc}
\toprule
\textbf{Model}
& \textbf{CKA($Z$, LowFreq)}
& \textbf{CKA($Z$, HighFreq)} \\
\midrule
Harmon
& 0.424 $\rightarrow$ \textbf{0.461} $(\uparrow 8.7\%)$
& 0.223 $\rightarrow$ \textbf{0.192} $(\downarrow 13.9\%)$ \\
Show-o2
& 0.453 $\rightarrow$ \textbf{0.486} $(\uparrow 7.3\%)$
& 0.252 $\rightarrow$ \textbf{0.212} $(\downarrow 15.9\%)$ \\
\bottomrule
\end{tabular}}
\end{table}

\subsubsection{Frequency-Decomposed Token Analysis}

To further understand what information is preserved or suppressed by InfoTok, we perform a frequency-decomposed CKA analysis on the learned visual tokens. Specifically, we decompose each image into low-frequency (Gaussian-blurred, capturing structure and layout) and high-frequency (residual, capturing texture and noise) components, and measure the CKA between visual token representations and each component before and after applying InfoTok.  As shown in Tab.~\ref{tab:cka_frequency_both}, InfoTok consistently increases the alignment of visual tokens with low-frequency image components while reducing their alignment with high-frequency components on both Harmon and Show-o2. These results suggest that InfoTok does not simply compress the token space indiscriminately, but selectively shifts it toward stable structural information and away from high-frequency details. This observation is complementary to Tab.~\ref{infotok_vae_mi}: while Tab.~\ref{infotok_vae_mi} characterizes InfoTok from the perspective of information regularization in the latent space, Tab.~\ref{tab:cka_frequency_both} provides representation-level evidence that such regulation promotes a more structured and reusable information allocation.

\section{Conclusion}

We presented InfoTok, an Information Bottleneck grounded framework for information regularized visual tokenization in shared token unified MLLMs.
Motivated by a capacity constrained view, InfoTok makes information allocation in the shared token space explicit by regulating the dependence among images, shared tokens, and multimodal outputs, and can be instantiated with practical estimators such as a variational formulation or an HSIC based measure.
Experiments on representative unified MLLMs show consistent improvements on both understanding and generation without introducing additional training data or architectural changes.
Ablations further suggest that the canonical IB terms already yield strong gains, while alignment and distillation mainly improve stability and provide additional benefits.
These results indicate that improving shared tokenization can be approached as regulating information flow, offering a principled complement to architecture driven tokenizer design.

{\small
\bibliographystyle{IEEEtran}
\bibliography{IEEEfull}
}

\end{document}